\documentclass[acmsmall]{acmart}

\newcommand{\ignore}[1]{{}}

\newcommand{\pre}{\textit{pre}}
\newcommand{\eff}{\textit{eff}}

\AtBeginDocument{%
  }

\setcopyright{acmcopyright}
\copyrightyear{2022}
\acmYear{2022}
\acmDOI{XXXXXXX.XXXXXXX}

\acmJournal{TIST}
\begin{document}

\title{Integrating AI Planning with Natural Language Processing: A Combination of Explicit and Tacit Knowledge}

\author{Kebing Jin}
\email{jinkb@mail2.sysu.edu.cn}
\author{Hankz Hankui Zhuo}
\authornote{Corresponding author}
\email{zhuohank@mail.sysu.edu.cn }
\affiliation{%
  \institution{School of Computer Science and Engineering, Sun Yat-sen University}
  \city{Guangzhou}
  \country{China}
  \postcode{510006}
}

% \affiliation{%

% \affiliation{%

% \affiliation{%

% \affiliation{%

% \affiliation{%

% \affiliation{%

% \affiliation{%

\renewcommand{\shortauthors}{Jin et al.}

\begin{abstract}
Natural language processing (NLP) aims at investigating the interactions between agents and humans, processing and analyzing large amounts of natural language data. Large-scale language models play an important role in current natural language processing. However, the challenges of explainability and complexity come along with the developments of language models. One way is to introduce logical relations and rules into natural language processing models, such as making use of Automated Planning. Automated planning (AI planning) focuses on building symbolic domain models and synthesizing plans to transit initial states to goals based on domain models. Recently, there have been plenty of works related to these two fields, which have the abilities to generate explicit knowledge, e.g., preconditions and effects of action models, and learn from tacit knowledge, e.g., neural models, respectively. Integrating AI planning and natural language processing effectively improves the communication between human and intelligent agents. This paper outlines the commons and relations between AI planning and natural language processing, argues that each of them can effectively impact on the other one by five areas: (1) planning-based text understanding, (2) planning-based natural language processing, (3) planning-based explainability, (4) text-based human-robot interaction, and (5) applications. We also explore some potential future issues between AI planning and natural language processing. To the best of our knowledge, this survey is the first work that addresses the deep connections between AI planning and Natural language processing. 
\end{abstract}
\begin{CCSXML}
<ccs2012>
   <concept>
       <concept_id>10010147.10010178.10010179</concept_id>
       <concept_desc>Computing methodologies~Natural language processing</concept_desc>
       <concept_significance>500</concept_significance>
       </concept>
   <concept>
       <concept_id>10010147.10010178.10010199</concept_id>
       <concept_desc>Computing methodologies~Planning and scheduling</concept_desc>
       <concept_significance>500</concept_significance>
       </concept>
   <concept>
       <concept_id>10010147.10010178.10010179.10003352</concept_id>
       <concept_desc>Computing methodologies~Information extraction</concept_desc>
       <concept_significance>300</concept_significance>
       </concept>
   <concept>
	   <concept_id>10010147.10010178.10010179.10010182</concept_id>
       <concept_desc>Computing methodologies~Natural language generation</concept_desc>
       <concept_significance>300</concept_significance>
       </concept>
 </ccs2012>
\end{CCSXML}

\ccsdesc[500]{Computing methodologies~Natural language processing}
\ccsdesc[500]{Computing methodologies~Planning and scheduling}
\ccsdesc[300]{Computing methodologies~Information extraction}
\ccsdesc[300]{Computing methodologies~Natural language generation}

\keywords{AI planning, Natural language processing, Natural language understanding, Human-robot interaction, Explainability}

\received{xxxx}
\received[revised]{xxxx}
\received[accepted]{xxxx}

\maketitle

\section{Introduction}
Natural language processing (NLP) aims at investigating the interactions between agents and humans, processing and analyzing large amounts of natural language data. %Language models are the core components of modern natural language processing. 
In recent years, for attaining better performance and handling large corpora, building large-scale language models is an inevitable trend in real applications \cite{li2021terapipe,narayanan2021efficient,zhang2021cpm}. Despite the success of language models in various domains, the explainability and complexity of language models have drawn intense research interests recently. In order to make models explainable and lightweight, integrating models with symbolic planning has been demonstrated effective in various NLP tasks. Symbolic planning (AI planning) is a branch of artificial intelligence that focuses on building symbolic domain models and synthesizing plans to transit initial states to goals based on domain models. The plans are typically for execution by intelligent agents, autonomous robots, and unmanned vehicles. Different from classical control and classification problems, the solutions are complex and must be discovered and optimized in multidimensional space. 
%AI Planning is able to reason about rules between events, build domain models, and synthesize plans to transit initial states to goals. Therefore, 
% AI planning is widely used in workflow management, autonomous robots, and unmanned vehicles \cite{schumacher2004uav,DBLP:conf/icra/FinnL17}. 
Generally, those approaches are mostly based on structured data, which has a well-defined structure and logically explainable to humans. 

%% 自然语言处理是。。。，他受到广泛研究。。。。语言模型是其重要核心。近年来，为了取得更好的效果，处理越来越大的数据集，大语言模型是当前的趋势。尽管其成功，。。。一种思路是引入逻辑，例如引入智能规划。

Compared with structured data used in AI planning, natural language descriptions are often complicated by omissions, inverted order, etc., resulting in difficulties in reasoning about language descriptions. It is thus often hard to directly train neural models to generate available and correct solutions, although deep learning has been widely used to handle unstructured data. Deep learning methods do well in acquiring knowledge from data, capturing implied rules, and expressing them by mathematical and neural models, which are tacit and unable to be directly shared with other humans and agents. Different from deep learning methods that aim to learn tacit knowledge, planning-based methods are better at capturing changes, formalizing them by rules, and generating valid plans when handling structured data. Rules are already codified, namely explicit knowledge, which can be clearly expressed and easily shared with others. Therefore, AI planning is one of the considerable steps to understand implied rules and build domain models from large amount of texts in natural language processing \cite{DBLP:conf/aaai/MeiBW16,DBLP:conf/ijcai/FengZK18}. %% 加几个引用
% 除了传统应用结构化数学的的规划，目前许多的方向都蕴含着planning的思想，比如非结构化的图片和文本，只要对逻辑的梳理就可以与planning进行结合。比如一些文本抽取动作。。。

\begin{figure}[!ht]
    \centering
    \includegraphics[width=0.8\textwidth]{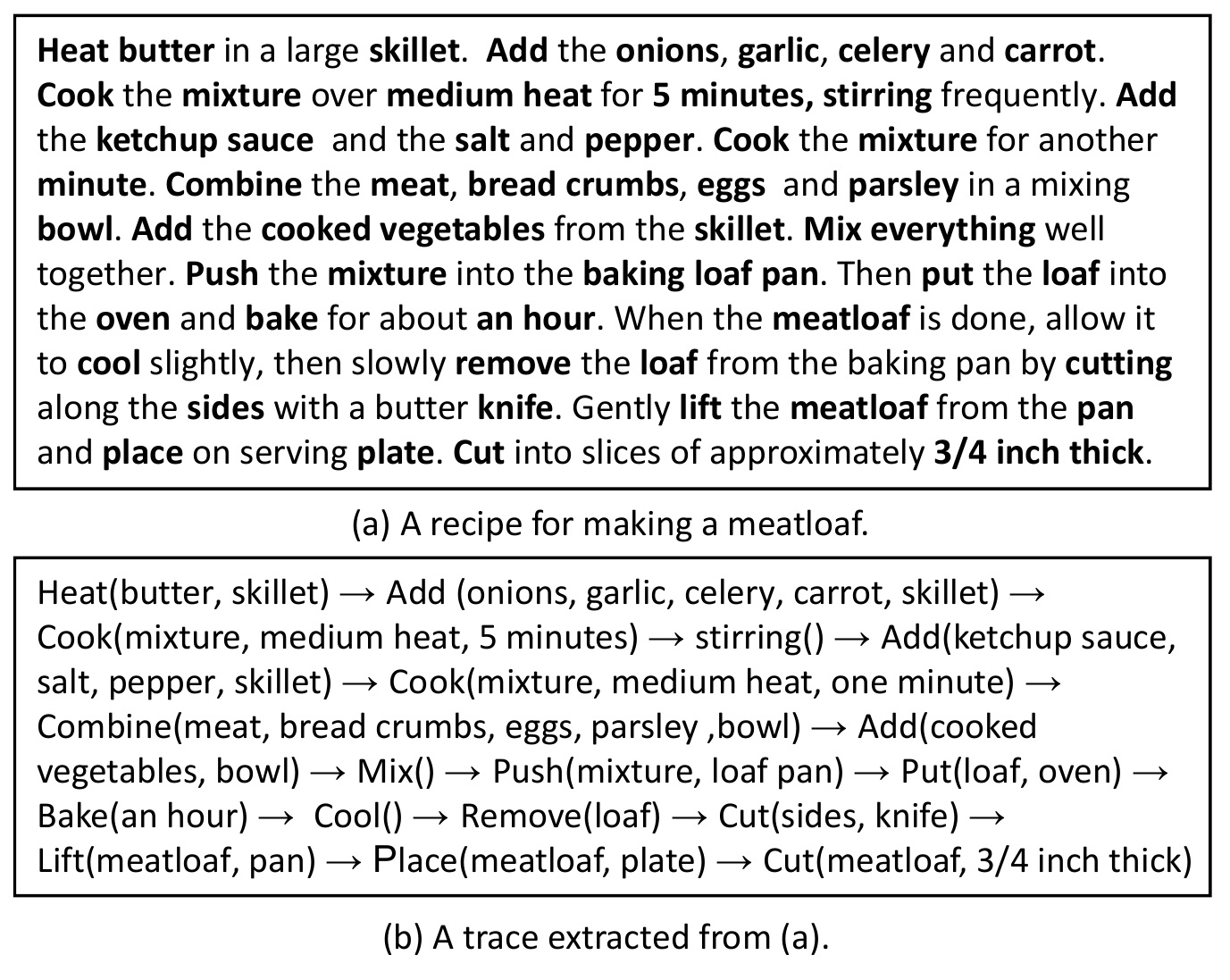}
    \caption{An example textual recipe about making a meatloaf.}
    \label{fig:example_introduction}
\end{figure}

On the other hand, unstructured data in real world is not disorderly but often a sequence based on rules. As for a natural language description, there is a theme running through it, along with a series of relevant events and a coherent text unfolds. Each sentence relates to the preceding texts and influences following sentences, just like preconditions and effects of actions in AI planning. \emph{For example, in a recipe about making a meatloaf shown in Figure \ref{fig:example_introduction}(a), humans can easily understand it and capture the main information including verbs, e.g., ``Heat'', and objects, e.g., ``butter'' and ``skillet''. However, as for agents, when given a mass of data in the form of sentences, it is hard to directly build models to reason about the implied rules and predict next moves. If we extract these information and formalize them structurally, as shown in Figure \ref{fig:example_introduction}(b), it is easier to construct models based on planning methods for guiding future unseen tasks. }

% 特别的，文本的描述就时常贯穿一个主题，每一句文本也都和上下文息息相关。比如指令性文本，里面的操作不是无中生有的，而是有着非常强的前提条件以及发生后需要执行的动作，比如。。。 因此学习动作模型。。。。

%% Figure example 文本需要逻辑

Besides using AI planning to help reason about implied rules in texts, the power of AI planning about capturing implied relations and computing valid solutions is another effective way to improve natural language processing, such as text summarization and machine translation. For example, there have been planning-based text generation methods \cite{DBLP:conf/aaai/YaoPWK0Y19,kong-etal-2021-stylized} extending a clear storyline ordered in advance. Those methods first compute sequences composed of keywords, key phrases, or contents as storylines, then use natural language processing techniques to extend storylines to coherent texts. \emph{In the above-mentioned example, generating an available recipe in a correct order shown in Figure \ref{fig:example_introduction}(a) is hard. However, given some rules, such as domain models about the operations of cooking, agents can compute plans toward achieving specified goals like a theme about making a meatloaf, as shown in Figure \ref{fig:example_introduction}(b). Agents can easily extend the plan and gain a valid recipe. }

% Both of planning-based natural language understanding and planning-based natural language processing mentioned above are productions of integrating AI planning and natural language processing, learning from tacit knowledge as well as using explicit knowledge. 

The integration of AI planning and natural language processing combines the best of tacit knowledge learning from sentences and explicit knowledge in the form of rules. As discussed in \cite{DBLP:journals/cacm/Kambhampati21}, it would be more effective to combine explicit and tacit knowledge rather than giving up explicit knowledge and learning everything from tacit knowledge, which is the current trend. Integrating AI planning and natural language processing allows human to communicate with agents in a more comfortable way, and enables intelligent agents to explain themselves to human in a human-understandable way. 
Natural language, as the most comfortable way to communicate with humans, establishes a relationship between humans and intelligent agents. In recent years, researchers have made efforts to connect with natural language and robots, such as by dialogue systems \cite{sreedharan2020d3wa+,pallagani2021generic} and natural language commands understanding \cite{kustenmacher2021improving,suddrey2022learning}. On the other hand, planning-based natural language models are based on structured data or implied rules, such as predicted storylines, which allows human to partly understand the principles of models. 

\begin{figure}[!ht]
    \centering
    \includegraphics[width=0.98\textwidth]{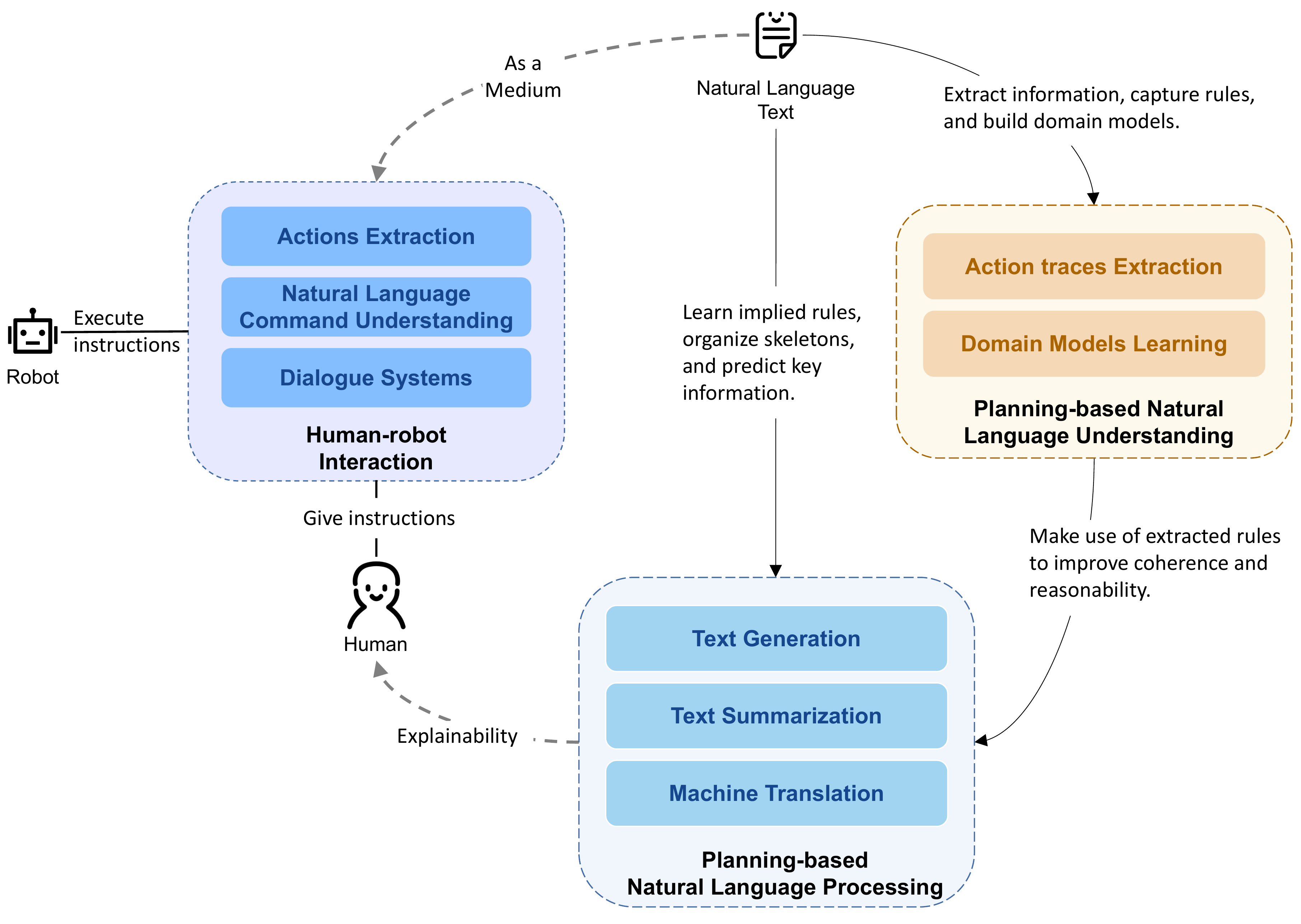}
    \caption{Relations between AI planning and Natural language processing}
    \label{fig:framework}
\end{figure}

In this paper, we first introduce some background knowledge in AI planning and natural language processing as well as their relations. Then we give a comprehensive overview of integrating AI planning and natural language processing by four aspects and their challenges: planning-based text understanding, planning-based natural language processing, planning-based explainability, and text-based human-robot interaction. Their relations are shown in Figure \ref{fig:framework}. Firstly, planning-based natural language understanding includes extracting actions from texts and learning domain models from texts. Secondly, we introduce planning-based natural language processing by three tasks integrated with AI planning, i.e., text generation, text summarization, and machine translation. Then we discuss planning-based explainability. Next, we introduce text-based human-robot interaction by extracting actions from natural language instructions, natural language command understanding, and dialogue generation. Finally, we present current applications, several future directions and conclude this paper. To the best of our knowledge, this survey is the first work that addresses the deep connections between AI planning and NLP. 

\section{Planning domain description language and NLP}

In this section, we introduce modeling knowledge in AI planning, backgrounds in natural language processing (NLP), and relations between AI planning and NLP, including similarities, differences, and language model-based planning. 

\subsection{Planning domain description language}
A planning problem is composed of a planning domain $\mathcal{D}$ and an instance $p$, defined by planning domain description language. With the development of AI planning, more and more extended planning domain description languages \cite{pddl3,pddl+,rddl} have been proposed. Taking PDDL (Planning Domain Definition Language) \cite{PDDL} as an example, a planning domain $\mathcal{D}$ is made up by several action models. An action model is defined by a tuple of $A = \langle a, \pre(a),$ $\eff(a)\rangle$, where $a$ is an action name with zero or more types of parameters. An action is a grounding of an action model, each of whose parameters is an object. $\pre(a)$ is a set of preconditions requiring to be satisfied when executing $a$, each of which is a proposition or a numeric constraint. Similarly, $\eff(a)$ is a set of effects, an effect can be a topical proposition, added into or deleted from the state after executing $a$, or a numeric updating, increasing or decreasing the value of variables according to specified functions. An instance is defined by $p = \langle s_0, g, \xi \rangle$, where $s_0$ is a set of initial assignments by propositions and variables, and $g$ is a set of goals requiring to be achieved. $\xi$ is an objective function guiding planner to compute for a minimum cost or maximum reward. A planning problem is to compute an available action sequence, which can transfer $s_0$ to a state containing desired goals $g$. 

\begin{figure*}[!ht]
    \centering
    \includegraphics[width=0.98\textwidth]{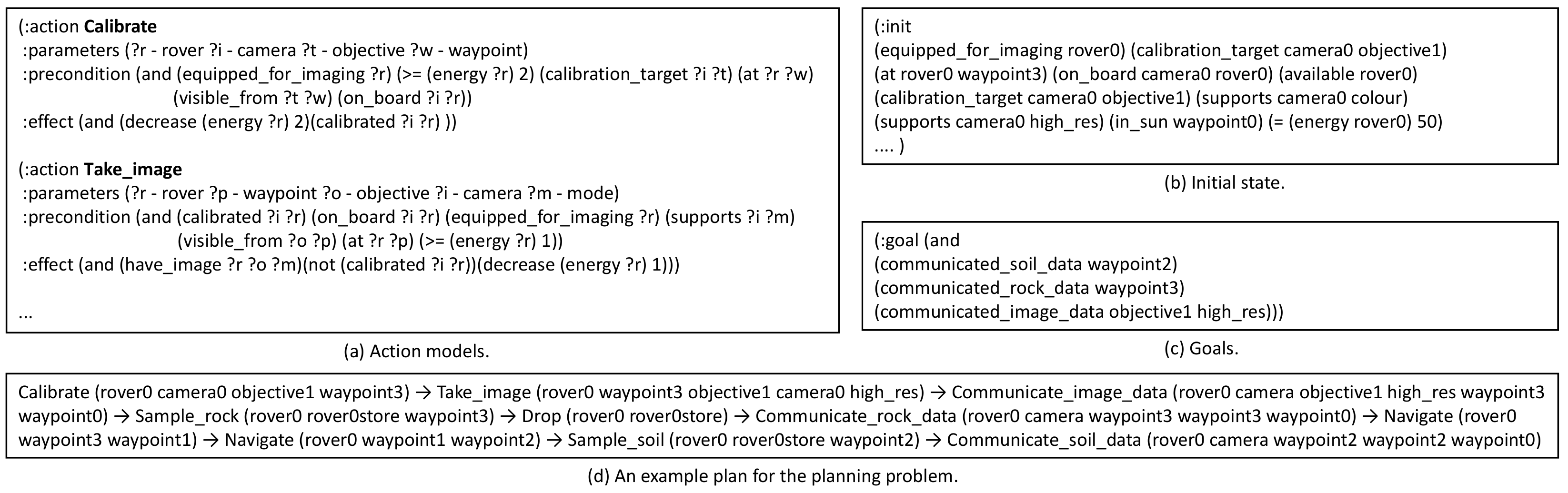}
    \caption{An example in the Rover domain, including action models, initial state, goals, and an available plan.}
    \label{fig:example_planning}
\end{figure*}

\emph{For example, parts of action models in the Rover domain are shown in Figure \ref{fig:example_planning}(a), where ``(equipped\_for \\
\_imaging ?r)'' is a proposition asking that a rover ``?r'' should equip with a camera for taking image when executing action ``Calibrate''. ``($>$= (energy ?r) 1)'' is a numeric precondition requiring the energy of rover ``?r'' should be larger than 1. Figure \ref{fig:example_planning}(b) and (c) show an initial state and goals, respectively. An example valid plan, updating the initial state to a state achieving the goals, is shown in Figure \ref{fig:example_planning}(d), where each action is a grounding action model with parameters. }

\subsection{Natural language processing}
Recently, natural language processing (NLP) \cite{cambria2014jumping,torfi2020natural,khurana2022natural} has attracted lots of attention, and it builds a bridge between human and agents. Natural language processing is grand, including various fields, such as natural language understanding (NLU)\cite{storks2019commonsense,otter2020survey}, natural language generation (NLG), \cite{gatt2018survey,dong2021survey} machine translation\cite{okpor2014machine}, and spelling correction\cite{hladek2020survey}. NLP has undergone several stages of rule-based models, statistic-based models, and neural network models. Rule-based NLP \cite{shaalan2010rule} is led by hand-crafted rule sets, whose main task is to understand natural language. It is, however, difficult and time-consuming to build all hand-crafted rules, lacking of scalability. Statistic-based NLP \cite{statisticMTsurvey} makes use of probability distributions to generate proper words and sentences, promoting the application of statistical machine learning methods based on large-scale corpora in natural language processing. Nevertheless, statistic-based models is barely to capture long-term relations and use information included in contexts. Recently, deep learning has been widely used in NLP tasks\cite{DBLP:journals/jair/Stahlberg20}, it is able to capture tacit knowledge implied in texts. However, deep learning is to fit neural networks and predict based on statistics, it can not ``understand'' the real meaning in natural language. In this paper, we focus on those NLP tasks related to AI planning. Compared with NLP approaches totally based on deep learning, planning-based NLP methods are more curious about implied logic and reasons of the solutions.

\subsection{Relations between AI planning and natural language processing}
In this section, we will sketch the relations between AI planning and natural language processing, which is mentioned by \cite{DBLP:journals/cogsci/Wilensky81,DBLP:conf/ijcai/GeibS07,young2013plans}. We will first introduce the similarities between AI planning and natural language processing, and then talk about their differences and deep relations. 

We discuss the similarities between AI planning and natural language processing by two aspects.
First of all, AI planning and natural language processing both revolve around observations and knowledge \cite{DBLP:conf/ijcai/GeibS07}. Planning tasks aim at either solving problems based on a current observation and goals along with priori knowledge like action models, or constructing knowledge such as transition functions based on sequential observations. Similarly, in natural language processing tasks, a text can be regarded as a sequence of observations, each observation is a sentence describing a partially observed state, where observations changes following implied rules. Secondly, they share two major common problems in planning tasks and natural language processing tasks: consistency and diversity. Both of plan traces and texts are cohesive descriptions and stringed by some rules and goals. As for planning, the rules are action models and transition functions, which update current states to next ones. The goals are sets of goal states or objective functions, guiding planners to attain optimal solutions. As for natural language processing, the rules are implied in the organization of texts, such as transition connections or consequences. Their goals can be titles, themes, or topics. On the other hand, in natural language processing, given goals, we can generate lots of coherent texts composed of different events, similar to different plan traces computed for the same goal states. Moreover, a sequence of events can be written as various stylized texts. As shown in Table \ref{table:planning_NLP}, we enumerate some concepts in AI planning and natural language processing, which can have some similarities. \emph{For example, objects, such as ``rover0'' and ``objective1'' in Figure \ref{fig:example_planning}, in planning problems are similar to entities in texts, e.g., ``skillet'' and ``onions'' in Figure \ref{fig:example_introduction}. }

\begin{table}[!h]
    \centering
    \caption{Similar concepts between AI planning and natural language processing}
    \setlength{\tabcolsep}{10mm}{
    \begin{tabular}{c|c}
    \toprule
        \textbf{AI planning} & \textbf{Natural language texts} \\
        \midrule
        Objects & Entities \\
        States & Sentences, sentiments, and intentions  \\
        Actions & Events \\
        Domain models & Implied rules, transitions, and relations  \\
        Goals & Topics, and themes \\
        Plan traces & Storylines, skeletons, and frameworks  \\
    \bottomrule
    \end{tabular}}
    \label{table:planning_NLP}
\end{table}
%%不同点 与 为什么要结合
%% (1) 文本任务需要知识作为基础，planning可以提供知识
%% (2) 文本任务需要多样性，planning在解决问题时通过调整搜索策略提供多样的结果
%% (3) 
% Besides the commons in planning and natural language processing, there are several relations between them. 

Although AI planning and natural language processing have those commons, the difference between them is that AI planning is good at generating explicit knowledge, such as domain models \cite{AMAN,DBLP:journals/ai/Zhuo014,DBLP:journals/ai/ZhuoM014,DBLP:journals/ai/ZhuoK17}, while natural language processing often learns tacit knowledge, such as training models from natural language data. \cite{DBLP:journals/cacm/Kambhampati21} argues that AI systems should be able to know when to take advice and when to learn, to find a balance between explicit and tacit knowledge. Taking planning-based text generation as an example, although texts are required to be coherent and with correct logic, natural language processing is weak in computing available and valid events, which AI planning is good at. And the integration of both allows agents to generate coherent texts by first using AI planning to generate storylines and then learning text generator by natural language processing techniques.

In a word, there are close ties between AI planning and natural language processing, due to their different advantages of explicit and tacit knowledge. The combination allows each of them to effectively impact on the other one.  

\subsection{Language Model-based Planning}

Although AI planning can effectively capture rules from action sequences, it is hard for humans without expert knowledge to construct structured plans. A natural and intuitive way is to use human natural language to describe plans, asking language model-based planners to be able to handle text sequences and predict the next moves \cite{DBLP:journals/tist/ZhuoZKT20,DBLP:conf/aaai/Say21}. Prior works mostly make use of pre-trained language models (LMs) to understand abstract, high-level textual actions and learn actionable knowledge for guiding planning \cite{DBLP:conf/emnlp/Jansen20,DBLP:journals/corr/abs-2202-01771,DBLP:journals/corr/abs-2207-05608}. Specifically, Huang et al. \cite{DBLP:conf/icml/HuangAPM22} use large language models (LLMs) to generate natural language actions, they investigated actionable knowledge already contained pre-trained LLMs. On the other hand, some works map natural language instructions and high-level goals to actions and goals, and learn policy to make decisions \cite{DBLP:conf/acl/Sharma0A22}. For example, LID \cite{DBLP:journals/corr/abs-2202-01771} uses policies initialized with pre-trained LMs and fine-tunes policies for predicting actions, validating that LMs are able to contain rich actionable knowledge. Language model-based planners are mostly based on templated textual actions datasets, rather than complex natural language instructions with various styles of descriptions. In the following sections, we introduce deep connections between AI planning and natural language processing, to rise to the dual challenges from rules and natural language.

\section{Planning-based natural language understanding}
%% 任务是干什么的
In this section, we introduce a comprehension overview of natural language understanding based on AI planning. Natural language understanding aims at comprehending human language, including sentiments, relations in contexts, topics, etc. Compared with learning relations from structured state traces, learning relations between extracted events from texts are even more challenging. It asks agents to reason about contexts, capture themes of sentences by selecting words to represent them, compute causal relationships between the selected words. There are two major areas introduced in the following section, which are both based on AI planning, to understand the texts and learn causal relationships from texts. The first one is to extract action sequences from texts, which requires agents to understand complex contexts from action descriptions. Moreover, it requires agents to be capable of reasoning about connotations in texts, such as exclusion relations and optional relations between actions. The other one is to first select words to represent the main ideas of sentences, and then learn the rules implied in sentences and formalize them by readable domain models for guiding agents to solve future unseen tasks and helping agents and people understand logical relations between events. 

\subsection{Extracting actions from texts}
\label{sec:extraction}
There have been works on extracting action sequences from action descriptions \cite{DBLP:journals/ijccc/ZhaoYLGL21}. The inputs of the task mostly include some texts describing some actions and procedures, the outputs are action sequences from the texts, each action is composed of a verb as action name and some objects. 
\begin{figure*}[!b]
    \centering
    \includegraphics[width=0.92\textwidth]{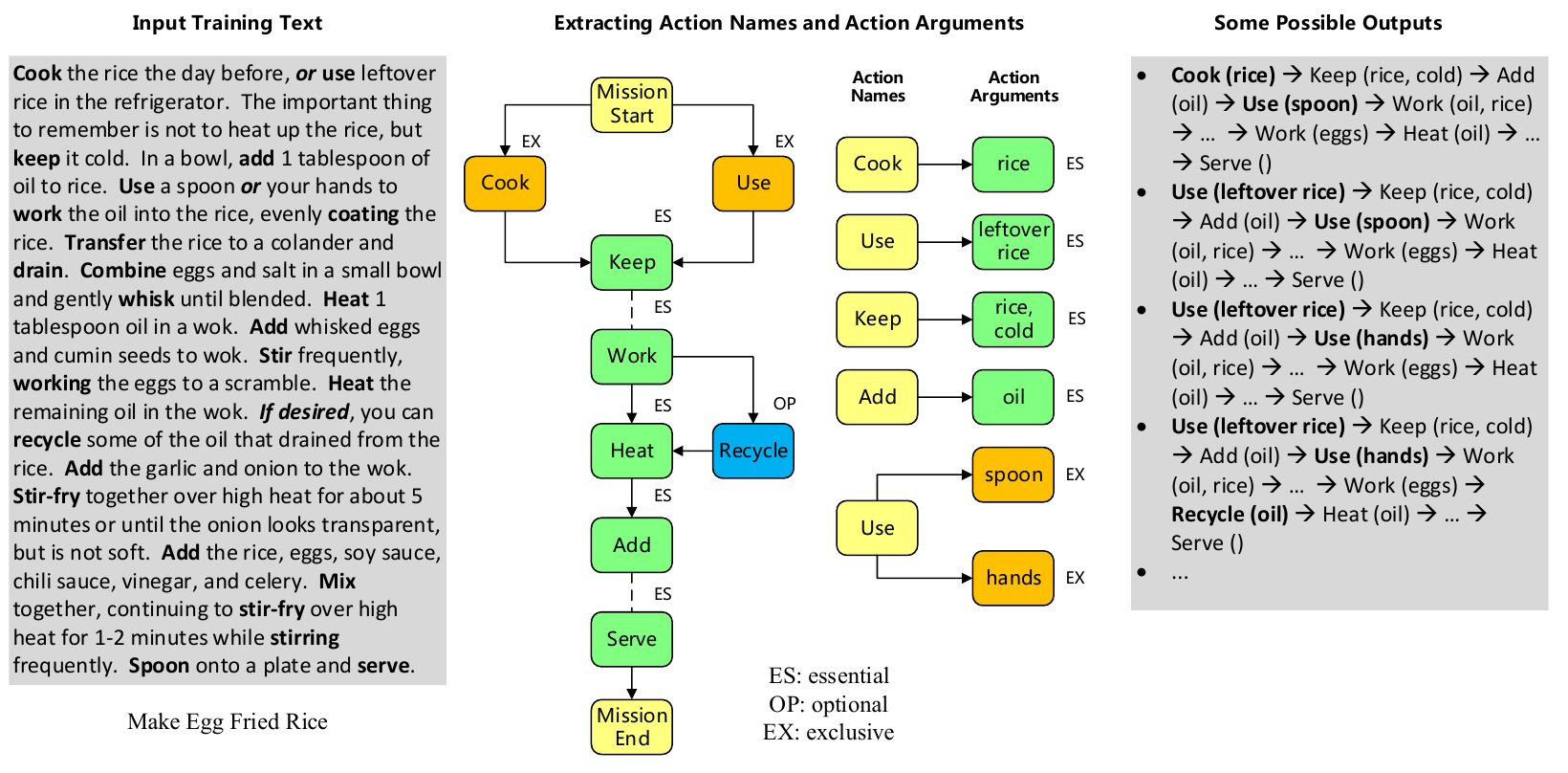}
    \caption{An illustration of action sequence extraction problem in \protect\cite{DBLP:conf/ijcai/FengZK18}}
    \label{fig:extracting}
\end{figure*}
Figure \ref{fig:extracting} shows an example in \cite{DBLP:conf/ijcai/FengZK18}, where an input text is in left part of Figure \ref{fig:extracting}, extracting action traces are shown in the right part of Figure \ref{fig:extracting}, the relations between actions are shown in the middle of Figure \ref{fig:extracting}. This task does not only need extract a word standing for an action of the sentence, but also reason about contexts for completing omissions caused by pronouns. Early approaches \cite{DBLP:conf/hri/MatuszekFK10,DBLP:conf/emnlp/KimM12} mostly make use of specialized resources, such as semantic parsers and learned lexicons, to reason about natural language route instructions. For example, MARCO \cite{DBLP:conf/aaai/MacMahonSK06} was proposed to map free-form natural language route instructions to action sequences, arising great interest in natural language processing community. MARCO is able to model a sentence by an instruction, e.g., ``Turn to face the green hallway'' can be modeled by ``Turn(until=(object=Path, appear=Green, side=Front, dist=0:))''. \cite{DBLP:conf/aaai/ChenM11} presented a system along with a plan refinement algorithm to transform natural language navigation instructions into executable formal plans. Generally, methods with semantics parsers require high simplicity of the texts. Therefore, mostly approaches are based on instructional texts or similar texts following some templates.

In recent years, learning methods, such as reinforcement learning and LSTM, have been widely used in natural language processing, as well as extracting action sequences from natural language texts, with the rapid development of artificial intelligence. 
For example, \cite{DBLP:conf/acl/BranavanCZB09} proposed a reinforcement learning approach for mapping natural language instructions in two domains, Windows troubleshooting guides and game tutorials, to sequences of executable actions. It uses a reward function to define the quality of the executed actions, and a policy gradient algorithm to estimate the parameters of a log-linear model for action selection. The learner repeatedly constructs action sequences for a set of documents, executes those actions, and observes the resulting reward. To handle free natural language without restricted templates, EASDRL \cite{DBLP:conf/ijcai/FengZK18} was presented to extract action sequences from texts, making use of deep reinforcement learning. It builds Q-networks to learn policies of extracting actions and extract plans from the labeled texts. EASDRL regards texts associated with actions as “states”, and associating words in texts with labels as “actions”. During capturing relations, EASDRL considers previously extracted actions as parts of states for deciding the choice of next operations. Therefore, EASDRL is able to reason about connotations in texts, such as exclusion relations and optional relations between actions. 
Except for reinforcement learning, there are more techniques used in extracting actions from texts. With the help of with long short-term memory (LSTM) recurrent neural networks, Mei at al. \cite{DBLP:conf/aaai/MeiBW16} proposed a neural sequence-to-sequence model to translate single-sentence natural language instructions to action sequences based upon a representation of the observable world state. LSTMs are applicable to a number of sequence learning problems, due to their ability to learn long-term dependencies, and they have been shown to be effective in tasks existing sequences. The LSTM framework allows agents to bidirectionally encode the navigational instruction sequence and decode the representation to an action sequence, based on a representation of the current state. 
% \cite{DBLP:journals/corr/abs-2106-07131} proposed a general purpose natural language model for action sequence extraction tasks using GPT-3, e Generative Pre-trained Transformer 3 (GPT-3) \cite{DBLP:conf/nips/BrownMRSKDNSSAA20}. 

%% 加讨论

% 2. extract actions from texts

\subsection{Learning domain models from texts}
%% what is PDDL 
Besides extracting action sequences from texts, another way to understand text is to learn the implied relations from sentences. The input of the learning task is a set of texts, and the output is a planning domain model composed of action models describing the relations by propositional preconditions and effects following the syntax of planning domain description language, such as PDDL \cite{PDDL}. Preconditions and effects make use of propositions to describe conditions that must be satisfied when executing actions and results after executing them, respectively. 

Learning domain models from instructional texts is a little different from narrative stories. 
Instructional texts are simpler than narrative stories, and the words are domain-dependent. Narrative stories are mostly third person synopses and they are always along with omission, which results in complexity when parsing sentences. 
% Therefore, previous researches for learning domain models from instructions focus on computing causal relationships from textual instructions rather than parsing sentences. 
A general way to construct a domain model is first to extract words and objects by parsing sentences for annotations (e.g., OpenNLP\footnote{https://opennlp.apache.org/} and Stanford CoreNLP\footnote{http://stanfordnlp.github.io/CoreNLP/}  ), and then learn causal relationships between them and formalize the relations by action models. 

To learn domain models from instructional texts, Sil and Yates \cite{DBLP:conf/ranlp/SilY11} used text mining via a search method to identify documents that contain words that represent target verbs or events. Then they used inductive learning techniques to identify appropriate action preconditions and effects. The method relies on handcrafted Pointwise Mutual Information to learn a SVM-based classifier that scores preconditions for a given action. Branavan et al. \cite{DBLP:conf/acl/BranavanKLB12} presented a reinforcement learning framework to extract precondition and effects relations implied by the text, and used these relations to compute action sequences for completing given tasks in the environment. Single argument predicates are extracted from the text as states, and regarded as sub-goals to construct hierarchical planning problems. Yordanova and Kirste \cite{DBLP:conf/icaart/YordanovaK16} extracted verbs and objects from text instructions based on part of speech (POS) tagging module, and discovered causal relations on the basis of the order of appearance to build PDDL models. However, due to lacking of connections between texts and world states and analyses between variable texts with the same meaning, it is hard to directly construct domain models as learning action models from structured data. In this paper, domain models are constructed according to some templates after parsing sentences. For example, ``If the apple is ripe, put the apple on the table. '' indicates that ``ripe'', a state of an apple, is a precondition of action ``put''. Therefore, a precondition of action ``put'' is ``(state-ripe)''. Although the model is readable for human, it is kind of redundant and not easy to understand for agents because of synonyms and polysemous words. Similarly, Lindsay et al. \cite{DBLP:conf/aips/LindsayRFHPG17} assumed texts are in restricted templates when describing actions. They generated sequences of actions by constructing representations of sentences and cluster operators by computing similarity, and built PDDL domain models with the help of a domain model acquisition tool. 
%These works focus on classifying events in instructions and summarizing the causality relationship between them. 

Considering the power of AI planning about offering correct causality and flexible narrative generation possibilities, constructing domain models has been used in narrative systems recently. Hayton et al. \cite{DBLP:conf/aaai/HaytonPFL20} proposed an approach taking natural language sentences which summarise the main elements of stories as inputs and generating action representations following PDDL, a narrative planning domain model. To overcome difficulties in parsing narrative stories, they presented two sets of rules to handle pronouns in stories. Then they used a template similar to \cite{DBLP:conf/icaart/YordanovaK16} to construct planning domains. Another specific difficulty for planning-based narrative systems is that hand-crafted domain models require more narrative actions and types of narrative objects compared to generated planning domains. The plentiful actions and objects let generated plan traces and storyline be more interesting and they can be extended to enjoyable stories. To achieve it, Porteous et al. \cite{DBLP:journals/aamas/PorteousFLC21} tried to anticipate the consequences of plan failure and the remedial actions or objects needed, or described several potential alternatives. They extended narrative planning domains by two types of principled mechanisms to operationalize narrative action and object substitution during narrative plan authoring. An original domain model can be extended with the addition generated by two mechanisms alternately. 

\subsection{Challenges and future prospects}
In AI planning, rules implied in states and actions are enforcedly constrained by preconditions and effects. Compared with AI planning, rules between events, such as concurrence, causality, and progression, in natural language processing are more flexible, which often intricately correlate with others. Moreover, the preconditions and effects are implied in natural language, which are abstract and hard to be modeled by propositions. In the other hand, in natural language text, a event can be written in different words, and a word owns various meanings. It therefore is difficult to directly distinguish them by only parsers. Those challenges make it hard to extract detailed logical relations implied in natural language, let along model implied relations by constructing structural domain models. It might be interesting to dig out clear logical relations implied in natural language text, which can lay a foundation for natural language processing tasks, such as explainability and controllable text generation.

%% 文本中的关系相比于规划中的关系（被前提条件与效果明确的限制）更为灵活，存在前后因果、并列、条件等关系，这种关系的捕捉具有挑战性。同时由于文本常用同义词或多意，单纯使用文本解析工具根据动词名词来捕捉是不可靠的。
%%这些关系难以捕捉且难以表达在plan中，更不用说以domain model来对其进行建模，但是interesting，因为可以更清晰的建立文本写作的脉络。

\section{Planning-based natural language processing}
In this section, we introduce three natural language tasks integrated with AI planning, including text generation, text summarization, and machine translation. Planning-based natural language processing tasks concern about reasonability and coherence, making use of the power of AI planning about reasoning about rules and relations.

\subsection{Planning-based text generation}
%% 文本生成任务是什么
% Recently, there have been significant advances in story generation. Text planning is one of crucial steps to guide models to generate well-organized long text. In general, prior studies with text planning usually first planning a sketch and then generate the whole story from the sketch. The sketch can be a sequence of keywords \cite{DBLP:conf/aaai/YaoPWK0Y19,kong-etal-2021-stylized,DBLP:conf/aaai/YuLCY021}, key phrases \cite{DBLP:conf/emnlp/XuRZZC018} or contents \cite{hua-etal-2021-dyploc}. These approaches were able to generate long texts with valid storylines. However, they cannot handle goal-driven tasks, because they cannot reason about the causal relationships between sentences.

One important field combined with AI planning is text generation, in which there have been significant advances, recently. Text generation asks models to generate coherent and interesting text based on preceding parts of the text, topics, titles, or themes, requiring agents to be capable of generating valid and clear logical frameworks. AI planning is one of crucial steps to guide models to generate well-organized long texts, owing to its power in learning domain models and computing solutions for goal-driven tasks. In this section, we introduce planning-based text generation methods with respect to the following two features: 
\begin{itemize}
    \item \textbf{Symbolic planning text generation} combines text generation with a classical planning framework, taking prior knowledge, e.g., domain models formalized by planning domain description language, as extra inputs. 
    \item \textbf{Neural planning text generation} are neural generators combined learning with a skeleton planning, to make up for the difficulties in building hand-crafted domain models.
    
    % \item \textbf{Text planning methods} Methods with a storyline planning, which may be a sequence of keywords, key phrases, or contents.
\end{itemize}

%% 给定domain文件生成
\subsubsection{Symbolic planning text generation}
%% 文献少

% \subsection{Symbolic text planning methods}
In order to generate coherent text with correct logic, it is natural to give agents some prior knowledge about basic rules between events. 
On the other hand, early rule-based researches \cite{DBLP:journals/coling/PerraultA80,cercone1986accessing,baud1993modelling} about natural language processing explore constructing representations of texts and combine with hand-crafted rules. It, however, is hard and tedious to enumerate all rules. Therefore, making use of AI planning to capture implied rules in the form of domain models with symbolic representations and compute proper skeletons is a natural way, which can overcome the difficulties of manually constructing rules \cite{DBLP:conf/aips/KollerH10,DBLP:journals/llc/Garoufi14,lukin2019narrative}.
%\cite{DBLP:conf/acl/KollerS07,DBLP:conf/acl/GaroufiK10,DBLP:conf/aips/KollerH10,DBLP:journals/llc/Garoufi14,lukin2019narrative}. 
For example, Porteous et al. \cite{DBLP:conf/icids/PorteousC09} proposed an approach injecting narrative control into plan generation through the use of PDDL \cite{PDDL} state trajectory constraints, to express narrative control information within the planning representation. They constructed constraint trees according to input domain models, and injected control into automatically generated narratives system. With the help of constraints, the approach decomposes problems into sets of smaller subproblems using the temporal orderings described by the constraints, and solves subproblems incrementally by a planner. Intentional Partial Order Causal Link (IPOCL) planning framework \cite{DBLP:journals/jair/RiedlY10_IPOCL} is an extension of classical planning, it aims at finding a sound and believable sequence of character actions that transforms an initial state into a state arriving goals. IPOCL does not only create causally sound plot progression, but also reasons about character intentionality by identifying possible character goals that explain their actions and creating plans that explain why those characters commit to their goals. Compared to IPOCL, CPOCL \cite{DBLP:conf/aiide/WareY11_CPOCL} preserves the conflicting
subplans without damaging the causal soundness of the overall story to generate interesting stories. CPOCL is an extension of IPOCL that explicitly captures how characters can thwart one another in pursuit of their goals, which is the essence of narrative conflicts. Making use of hand-crafted, well-defined domain models, symbolic planning text generation methods have the ability to produce impressive results in limited domains.

\subsubsection{Neural planning text generation}
However, it is often tedious or difficult to build domain models by hand due to the high requirements of manual efforts and domain knowledge. Automatically learning domains and constructing storylines have significantly attracted researchers' attention recently \cite{porteous2021automated,simon2022tattletale}. 

To automatically learn domain models for helping generate coherent and valid stories, Li et al. \cite{DBLP:conf/aaai/LiLJR13} used a crowd-sourced corpus of stories to learn plot graphs that can then be used as constrained search spaces for sequences of story events, instead of relying on priori domain models. Specifically, the approach crowdsources a corpus of narrative examples of a new domain, automatically constructs domain models capturing different possible, non-contradictory story trajectories, and samples from the space of stories allowed by the domain model according to some story quality criteria. During the plot graph learning, learning mutual exclusion relations and optional events lets the generated story be coherent. C2PO \cite{DBLP:conf/aaai/AmmanabroluCBR21} learns a branching story graph structure that can be searched, and introduces soft causal relations as causal relations inferred from commonsense reasoning. It creates a branching space of possible story continuations that bridge between plot points that are automatically extracted from existing natural language plot summaries. 

Another way for constructing valid line of text is to construct storylines in advance, which can be skeletons, or sequences of keywords, key phrases, or contents. Xu et al. \cite{DBLP:conf/emnlp/XuRZZC018} generated skeletons composed of phrases learned by a reinforcement learning method, and then expanded skeletons to complete and fluent sentences. Fan et al. \cite{DBLP:conf/acl/FanLD19} proposed a novel approach which first generates plans in the form of predicate-argument structures, then generates stories with placeholder tokens to indicate entities, and finally replaces tokens by entities based on the global story contexts. The inputs of the task are short descriptions of scenes or events, and the approach outputs relevant narrative stories following the inputs.

Instead of generating skeletons with detailed prompts, some approaches first plan out storylines, which enable them to generate controllable stories with goals. \cite{DBLP:conf/aaai/YaoPWK0Y19} proposed a hierarchical generation framework that first planned a controllable storyline composed of keywords towards a goal (i.e., a title), and then generated a story based on the storyline. The RAKE algorithm \cite{rose2010automatic} takes each sentence as an input and combines several word frequency based and graph-based metrics to weight the importance of the words. The approach regards the most important word as the keyword. The storyline is planned out based on the title, previously generated sentences, and the previous keywords in the storyline. In the experiments, they explore two strategies, dynamic schema and static schema. Results show the static schema performs better than the other one because it plans the storyline holistically, thus tends to generate more coherent and relevant stories. Similarly, \cite{kong-etal-2021-stylized} made use of a related framework, which first plans out storyline composed of a sequence of keywords and then generates the whole story, to handle stylized story generation. Stylized story generation is to generate stories with specified style given a leading context. Keywords are selected following some emotion-driven style, such as ``fear'', ``anger'', and ``surprsie''. According to the stylized keywords, the approach can generate generates the whole stylized story with the guidance of the keywords. Yu et al. \cite{DBLP:conf/aaai/YuLCY021} followed Yao et al. \cite{DBLP:conf/aaai/YaoPWK0Y19} and used the RAKE algorithm to extract keywords to train a generation model for conducting keyword planning given story titles as inputs. Then the approach combines the story titles and the corresponding keywords of each story as the inputs of the graph module to automatically generate a graph for each story. According to the keywords, titles, story graphs, the approach encodes them into latent variables and further decodes them to generate the corresponding stories.

Except for generating stories according to prompts or goals, DYPLOC \cite{hua-etal-2021-dyploc}, a dynamic planning generation framework, takes a set of content items as inputs, each content item consists of a title, a set of entities, and a set of core concepts. To organize unordered content items, DYPLOC introduces a plan scoring network, which learns to dynamically select and order contents based on what has been produced previously while generating the outputs. 

\subsubsection{Challenges and future prospects}
Recently, text generation \cite{iqbal2020survey,jin2020recent,yu2022survey} has gained lots of attraction, aiming at letting intelligent agents express like humans. Numbers of methods to generate coherent texts have been proposed in recent years. Generating logical and controllable texts, however, is still a challenging task. AI planning is one of critical ways to enable agents to generate logical and controllable texts. Compared with symbolic planning text generation methods and neural planning text generation methods, the former is more capable of generating logical storylines with goals, and the texts generated by the latter are more diverse and coherent. Specially, symbolic planning methods can generate more explainable storylines, which is still challenging for deep learning methods. In general, it would be interesting to combine both for generating logical, controllable, and coherent text with diversity. Although there have been some approaches \cite{DBLP:conf/aaai/ShaMLPLCS18,DBLP:conf/acl/FanLD19,DBLP:conf/ijcai/TambwekarDMMHR19}  %\cite{DBLP:conf/aaai/MartinAWHSHR18,DBLP:conf/aaai/ShaMLPLCS18,DBLP:conf/acl/FanLD19,DBLP:conf/ijcai/TambwekarDMMHR19} 
proposed, based on the syntax of plans or representation of structure in AI planning, they use neural networks to predict unseen events instead of speculating based on logical relations implied. We hold the opinion that, compared with only learning blackbox neural models with implicit rules, appropriately combined with explicit logical relations would be a new attempt, which maybe contribute to natural language processing tasks.

%% 文本生成任务虽然已经有很多任务，但是生成有逻辑的或是可控制的文本依然非常具有挑战。符号文本生成的任务能够抓住文本中的逻辑、生成可控的故事线。神经文本生成能够生成多样、流畅的文本，但是难以解释故事线生辰的原因。如果能将两者结合，能在保证文本逻辑的同时生成多样且流畅的文本。
\subsection{Planning-based Text Summarization}
%% 文本摘要是一个很重要的任务，是干什么的，为什么重要，分为生成式与抽取式，其中与规划结合较多的为生成式
%% 文本摘要需要对文本的理解，并精炼出详细的信息，因此尝尝与内容规划结合，用以理解文本的内容并且根据脉络与主题生成摘要。例如。。。。
Text summarization is to extract important information from texts and generate new texts based on those information in the form of summarizes. It requires agents to understand texts, filter information from abundant descriptions and organize them to form summarizes, which is one of the most researched areas among the NLP community. Text summarization can be categorized into extractive and abstractive techniques. Extractive summarization aims at selecting subsets of words or sentences from input articles to summarize them. Abstractive summarization takes articles as inputs, tries to understand the texts and generate summarizes. In recent years, some researchers try to integrate text summarization models with AI planning. The combination of AI planning and text summarization is mostly based on deep learning abstractive summarization, making use of content planning which describes or predicts skeletons of articles. Planning-based text summarization methods first plan out skeletons of summarizies or compute probability distributions, and then generate the whole sentences based on the skeletons or predictions. For example, 
Narayan et al. \cite{narayan2021planning} first computed plans in the form of entity chains, which are ordered sequences of entities, and then generated summaries conditioned on the plans. 
Marfurt et al. \cite{marfurt2021sentence} proposed an abstractive summarization model implemented with a planning step, done by a hierarchical decoder, which first plans out an outline for the next sentence in the form of sentence representations and generates words according to the representations. 
Amplayo et al. \cite{DBLP:conf/aaai/AmplayoAL21} incorporated content planning in unsupervised summarization and datasets creation. They predicted aspect and sentiment probability distributions as content plans and generated sentences according to the predictions. During creating datasets, they made use of the distributions parametrized by the content planner to control the structures of created datasets. 

In current planning-based text summarization approaches, content planning is based on deep learning which learns models and probability distributions to predict. However, only fitting deep learning models based on representations of sentences and words is hard to capture the relations implied. We believe that it would be interesting and challenging to let content planning modules understand relations implied in texts, which will allow content plans to be more coherent and controllable. On the other hand, although some mainstream text summarization methods are not based on AI planning, there are some commonalities between them. For example, tree-based and graph-based text summarization approaches\cite{DBLP:conf/ijcai/BanerjeeMS15,kurisinkel2017abstractive,DBLP:journals/jbi/AzadaniGD18} first find the most important information from the text and then use trees and graphs to create summaries. Those structures aim at representing the relations between sentences, which is similar to the relations between actions in AI planning but at a more abstract level. Secondly, some text summarization methods try to obtain important words from sentences, such as verbs, objects, and subjects, to represent sentences semantically \cite{alshaina2017multi}. These forms have similarities with the structured representations in AI planning such as propositions and first-order predicates. Thirdly, ontology-based text summarization methods \cite{DBLP:conf/emnlp/XiangJCS15,mohan2016study} collect entities and their relationships, which reminds us of domain models in AI planning. We believe that there is a vast scope for researchers to combine AI planning and text summarization.

%% challenges：除与内容规划的文本摘要任务以外，文本摘要已有许多基于文本结构的任务，比如predicate-argument-based，graph-based的方法。这种表示方式与规划的结构化表达相当相似，另一方面也有基于文本的领域知识的方法，ontology-based。这与规划的domain模型用途一致，用于表达domain中的实体与关系。这些可以与规划方法结合

\subsection{Planning-based Machine Translation}

% 近年来，神经机器翻译模型获得了极大的关注，其中一个方向是将其与规划所结合，组织排序所要生成的内容，再扩展为翻译的target language。
Machine translation aims at automatically translating content from source language to another target language, having a long history. One way to understand source texts and generate target texts is to combine neural language models with planning phases, i.e., first generating skeletons, and extending them by target languages. For example,  G{\"{u}}l{\c{c}}ehre et al. \cite{DBLP:conf/rep4nlp/GulcehreDTB17} integrated an auto-encoder with a planning mechanism, the auto-encoder first encodes texts by sequences of vector representations and decodes representations by generating target translation character-by-character. Specifically, they first created plans ahead in the form of action matrices, which are sequences of probability distributions, and made use of commitment plan vectors to govern whether to recompute plans or use them. Then they computed soft alignments based on the plan and generated texts in target language at each time-step. Shu and Nakayama \cite{DBLP:journals/corr/abs-1808-04525} combined neural machine translation with a planning phase, which first generates planner codes to disambiguate uncertain information about the sentence structure and control the structure of output sentences. Bahdanau et al. \cite{DBLP:conf/iclr/BahdanauBXGLPCB17} used actor-critic methods from reinforcement learning (RL) to generate sequences. They showed that sequences can be used in machine translation tasks, gaining better translation performance. Those approaches first produce big pictures of output texts by planning, then generate complete sentences conditioned on plans. They take advantages of capturing implied rules to generate more accurate and coherent target texts.

% challenges: 文本中每一句话之间存在逻辑关系，机器翻译不仅要理解原文本中词之间、句子之间的关系，还要组织生成的目标语言。若要生成流利的目标语言，不能仅仅依靠word-by-word的生成。但是，如何让智能体理解文本的起承转合并将它对应到目标语言是非常有挑战性的。
In natural language texts, the transitions between sentences imply relations and rules. Machine translation tasks do not only need to understand word-level structures of sentences, but also need to capture sentence-level relationships. Only relying on word-by-word text generation is hard and challenging to generate coherent and logical texts, especially when generating accurate transitions between sentences. Current approaches to capture implied rules and generate plans are mostly based on neural black-box models, lacking the explainability of making decisions. Previous researches about rule-based translation are explainable, they are based on hand-crafted rules. Although those rules are explicit and accurate, it is hard to manually write rules and the rules are not scalable. Producing rules are time-consuming and tedious. However, it would be interesting if we regard it as action models learning in planning community, which structurally formalize rules by preconditions and effects. We believe that combining rule-based machine translation with planning and neural machine translation may spark new ideas.

% \section{Text-based explainable planning}
\section{Planning-based explainability}

\begin{figure}[!ht]
    \centering
    \includegraphics[width=0.7\textwidth]{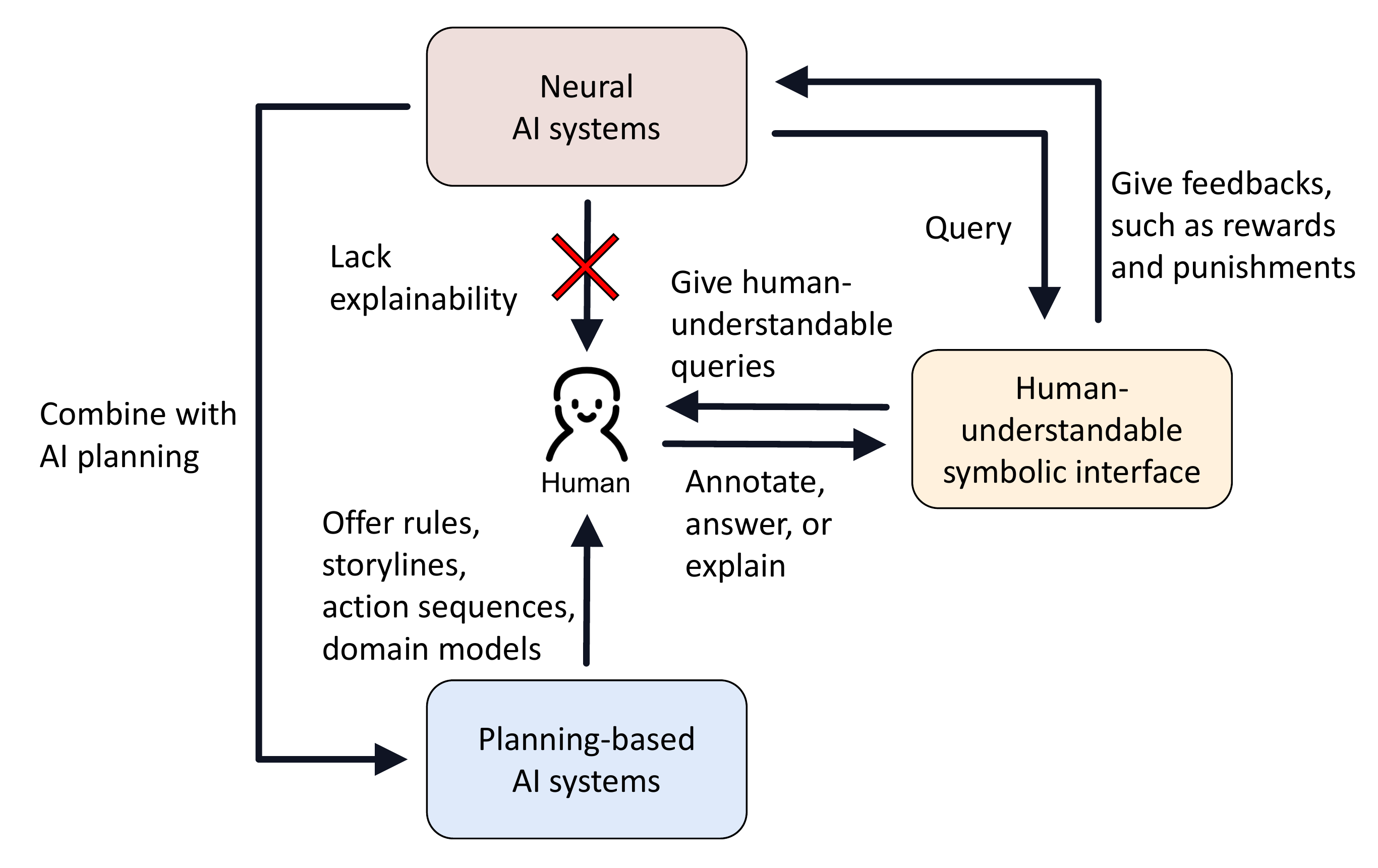}
    \caption{Two types of making neural AI systems be explainable: (1) integrating AI systems with planning techniques; (2) making use of human-understandable interface.  }
    \label{fig:explainability}
\end{figure}

Nowadays, although deep learning approaches have been widely used in AI fields, human cannot understand practical meaning inside black-box neural models. Differently, AI planning is able to offer explicit knowledge, in the form of first-order logic, domain models, etc, implied in natural language. Therefore, the combination of natural language and AI planning may enable AI systems to be able to explain their reasoning to humans, which meets the needs for AI systems to work synergistically with humans. Those systems require agents to be aware of the intentions, capabilities and mental model of the human in the loop during its decision process. As shown in Figure \ref{fig:explainability}, besides extracting action sequences \cite{DBLP:conf/aaai/MacMahonSK06,DBLP:conf/ijcai/FengZK18} and building domain models \cite{DBLP:conf/acl/BranavanKLB12,DBLP:conf/aips/LindsayRFHPG17} mentioned above, another way for making AI system explainable is to construct human-understandable symbolic interfaces (cf.  \cite{DBLP:journals/corr/abs-2109-09904}). A human-understandable symbolic interface is not only developed for its own computational efficiency, but also beneficial to humans. EXPAND system \cite{guan2021widening} and SERLfD framework \cite{DBLP:journals/corr/abs-2110-05286}, respectively. EXPAND system accelerates Human-in-the-Loop deep reinforcement learning by using human evaluative feedback and visual explanation. SERLfD uses self-explanation to recognize valuable high-level relational features as an interpretation of why a successful trajectory is successful, allowing SERLfD to guide itself and improve the efficiency.

\section{Text-based human-robot interaction}
\begin{figure}[!ht]
    \centering
    \includegraphics[width=1\textwidth]{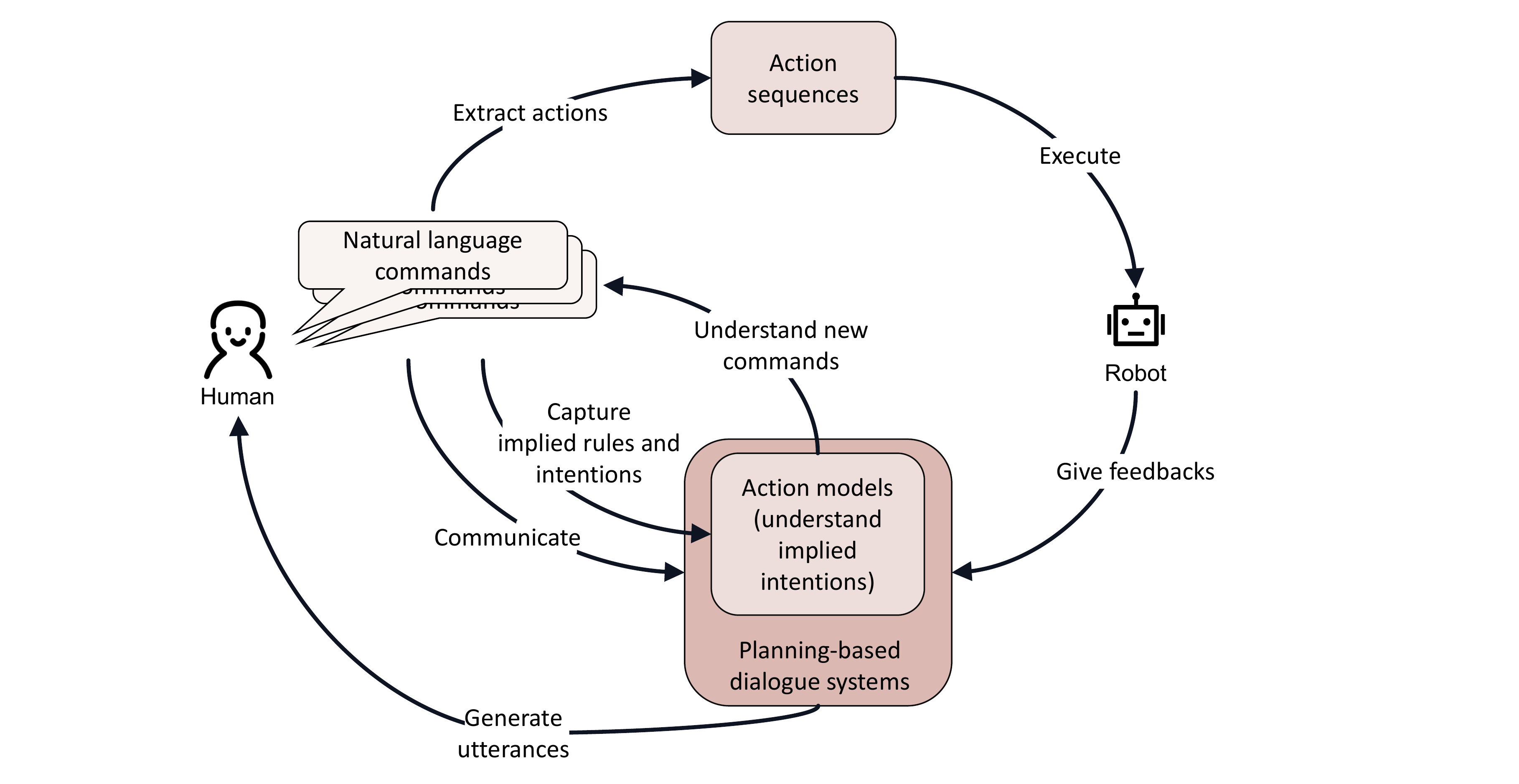}
    \caption{Relations of researches in human-robot interaction.}
    \label{fig:interaction}
\end{figure}

The rapid developments of artificial intelligence let robots move out from industrial environments, and enter the daily life of humans, such as homes and hospitals. It requires robots to be able to respond quickly and effectively to rapidly-changing conditions and expectations. Language–based communication is the most natural method for humans to communicate with others, so natural language is a good candidate to be robot instruction for human-robot interaction. In this section, we will introduce text-based human-robot interaction from three aspects: extracting actions from natural language commands, natural language command understanding, and dialogue generation, as shown in the Figure \ref{fig:interaction}.

\subsection{Extracting actions from natural language command}
One of important tasks in text-based human-robot interaction is to extract actions from natural language commands \cite{DBLP:conf/icra/TenorthNB10,nyga2012everything,pham2015extraction,hung2016extracting}. Extracting actions from natural language commands is similar to action extraction in Section \ref{sec:extraction}. Differently, extracting actions from natural language commands is not only to capture important words to indicate sentences, but also to understand implied rules and generate action sequences to guide robots to achieve tasks. Some approaches make use of with language descriptions, then build models that map language commands to action sequences. Tellex et al. \cite{DBLP:conf/aaai/TellexKDWBTR11} introduced a system, Generalized Grounding Graphs (G$^3$), taking a natural language command as input and outputting a plan for the robot. G$^3$ instantiates a probabilistic graphical model for a particular natural language command according to hierarchical and compositional semantic structure of the command. Cantrell et al. \cite{DBLP:conf/hri/CantrellTSBKS12} presented a robotic architecture equipping with a planner that uses newly discovered information to produce new and updated plans, specifically information originating in spoken input produced by human operators. The robot can learn action sequences with defined preconditions and effects from natural language descriptions, and immediately apply this knowledge to improve planning. On the other hand, some approaches \cite{DBLP:conf/indin/Buzhinsky19,DBLP:conf/robio/YeXLHSCS21} focus on temporal logic between natural language commands, aiming at handling semantic disambiguation of natural language.

\subsection{Natural Language Command Understanding}

Another important task is to enable agents to understand natural language commands given by humans, which requires agents to understand natural language commands and capture implied rules \cite{DBLP:conf/aaai/BiskSCM18,DBLP:conf/icra/ChenTKBA20,DBLP:conf/acl/SheC17,DBLP:conf/corl/BlukisMKA18,DBLP:conf/aaai/BiskSCM18,DBLP:conf/icra/ChenTKBA20}, or learn new actions based on natural language commands and dialogues\cite{DBLP:conf/sigdial/SheYCJCX14,DBLP:conf/aaai/MohanL14,DBLP:conf/atal/ScheutzKOFP17,DBLP:conf/aaai/FrascaOCS21}. To understand natural language commands, 
 %% 理解指令，抽取指令，理解指令，输出下一步的指令
Thomason et al. \cite{DBLP:conf/ijcai/ThomasonZMS15} introduced a dialog agent to understand human natural language commands through semantic parsing, actively resolve ambiguities using a dialog manager, and incrementally learn from human-robot conversations. The agent employs incremental learning of a semantic parser from conversations on a mobile robot. It is implemented and tested both on a web interface with hundreds of users and on a mobile robot over several days, tasked with understanding navigation and delivery requests through natural language in an office environment.
Brawer et al. \cite{DBLP:conf/iros/BrawerMRWS18} presented a framework for effectively grounding situated and natural language to action selection during human-robot interaction. It integrates verbal commands from a human partner with contextual information in the form of a task model. The approach is capable of acquiring and deploying new task representations from limited and natural language data sets, and without any prior domain knowledge of language or the task itself. 
Moreover, understanding action models and predicting next actions are widely researched for robotic tasks and navigation tasks with natural language commands \cite{DBLP:conf/corl/ThomasonMCZ19,DBLP:conf/cvpr/ChenSMSA19,DBLP:conf/aaai/ChiSEKH20}.

On the other hand, another challenge in natural language command understanding tasks is to learn new actions when facing unknown natural language commands \cite{DBLP:conf/ro-man/SheCCJYX14,DBLP:journals/aamas/AzariaSKLM20}. 
 %% 从交互中学习新的动作
Cantrell et al. \cite{DBLP:conf/ro-man/CantrellSS11} introduced an algorithm, to learn meanings of action verbs through dialogue-based natural language descriptions and integrated it in the robot’s natural language subsystem. The algorithm allows robots to perform the actions associated with the learned verb meanings right away without any additional help or learning trials. Moreover, it allows human to interact with a robot to explain new action words in natural language, and lets the robot be able to perform the new action and store the procedural knowledge for future usage. Learning by Instruction Agent (LIA) \cite{DBLP:conf/aaai/AzariaKM16} was proposed to learn new commands by natural language interaction with human. When facing a new natural language command that LIA does not understand, it prompts users to explain how to achieve the command through a sequence of natural language steps. LIA interprets commands using a semantic parser that maps each command to a logical form, which contains one or more of functions and predicates. 
 
Undoubtedly, letting agents understand natural language commands and infer about next actions is an important but challenging task. An agent does not only need to capture rules implied in utterances, but also need to distinguish sentences described in different ways. We believe that the combination of planning-based natural language processing and human-robot interaction would create something interesting for natural language commands understanding tasks.

\subsection{Dialogue Systems}
Dialogue systems \cite{DBLP:journals/sigkdd/ChenLYT17,DBLP:journals/tois/MaLZLL22} have been a bridge between human and robots, which interacts with human in natural language. AI planning is one of crucial mechanisms used in dialogue systems to recognize the intentions conveyed in dialogues \cite{camilleri2002dialogue,muise2019planning}. Planning-based dialogue systems take advantage of the power of capturing and expressing rules and use it to manage utterances or guide the generation. Rules, plans and intentions offer proper logical forms which derive appropriate communication acts in dialogue systems \cite{cohen2020back}. Plan-based model %\cite{chu1994plan,grosz1996collaborative,bohus2009ravenclaw,galescu2018cogent,santos2022review} 
\cite{chu1994plan,santos2022review} were proposed to manage the intentions and information implied in dialogues. Those models describe the common activities and relations between utterances, and can be used in the following generation processes. However, planning-based dialogue systems are still in early stages \cite{petrick2016using,cohen2020back}, which facing the challenges of complex representations in open-domain, difficulties of manually constructing models and limitations of scalability of dialogue models. Nevertheless, we believe that planning could be a strong suit for dialogue systems by integrating with automatically domain models learning and searching strategies, especially for controlled dialogue generation. Moreover, we are interested in the explainability of planning-based dialogue systems, it would be interesting to know the reason for intentions generation.

\section{Applications}
In this section, we introduce some reality applications based on combinations of AI planning and natural language processing. AI planning is widely used in reality management systems, such as logistics management, workshop schedule, and reservoir operation. Moreover, natural language processing (NLP) helps human communicate with agents, the combination of AI planning and NLP enables applications to be found in many fields, such as emergency managements \cite{fogli2013knowledge,blindheim2020risk,costa2022survey,costa2022survey} and urban planning \cite{resch2016citizen,lieven2021enabling,lieven2021enabling}. For example, the Urban Redevelopment Authority (URA) Centre in Singapore \footnote{https://www.ura.gov.sg/Corporate/Resources/Ideas-and-Trends/AI-in-Urban-Planning} deployed Robotic Process Automation and NLP to help conduct operations for resource optimization. The combination allows routine tasks to make use of AI planning and NLP, such as chatbots for public queries, which capture information from large datasets, analyze textual feedback, make planning decisions, and respond intelligently. On the other hand, the navigation systems in daily use, such as Baidu Maps \footnote{https://map.baidu.com/} and Google Maps \footnote{https://www.google.de/maps}, combine making decisions and NLP techniques, which plan out routes with different objectives according to goals, and generate natural language suggestions to guide human. Moreover, agents learn from human commands and navigation datasets, helping agents understand human behaviors \cite{DBLP:conf/kdd/HuangWFZSL20,miah2022social}. 

\section{Conclusion}
In this paper, 
% we address various aspects of planning-based natural language processing methods with respect to planning-based text understanding and planning-based text generation. 
we consider that AI planning and natural language processing have strong ties, and we introduce recent works about four related tasks, i.e., planning-based text understanding, planning-based natural language processing, planning-based explainability, and text-based human-robot interaction. We first introduce backgrounds about AI planning and natural language processing and discuss commons between them, as well as their abilities to generate explicit knowledge, e.g., domain models, and learning from tacit knowledge, e.g., neural models. We then introduce methods of planning-based text understanding by extracting action sequences from texts and learning domain models from texts. Next, we give an overview of planning-based natural language processing about text generation, text summarization, and machine translation. Then, we introduce recent works in planning-based explainability and text-based human-robot interaction.

With this paper, we aim to provide a high-level view of AI planning and natural language processing for further studies, about integrating them for a combination of explicit and tacit knowledge. Combining learning from tacit knowledge and using explicit knowledge in a fully principled way is an open problem, although there are non-negligible relations between AI planning and natural language processing, allowing each of them can effectively impact the other one. However, there is not enough communication between these two fields. While many advances have been made in natural language processing by using AI planning algorithms, a significant amount of research is still required to understand the implied knowledge hidden in texts. Meanwhile, improving the ability to describe environments by domain models and solve large-scale planning problems is also beneficial to understanding texts and generating coherent and interesting texts. We believe that integrating AI planning and natural language processing, a complex combination of explicit and tacit knowledge, is a promising research area, which can improve the communication between human and intelligent agents.

\bibliographystyle{ACM-Reference-Format}
\bibliography{acmart}


\begin{thebibliography}{132}

%%% the Web address.

\ifx \showCODEN    \undefined \def \showCODEN     #1{\unskip}     \fi
\ifx \showDOI      \undefined \def \showDOI       #1{#1}\fi
\ifx \showISBNx    \undefined \def \showISBNx     #1{\unskip}     \fi
\ifx \showISBNxiii \undefined \def \showISBNxiii  #1{\unskip}     \fi
\ifx \showISSN     \undefined \def \showISSN      #1{\unskip}     \fi
\ifx \showLCCN     \undefined \def \showLCCN      #1{\unskip}     \fi
\ifx \shownote     \undefined \def \shownote      #1{#1}          \fi
\ifx \showarticletitle \undefined \def \showarticletitle #1{#1}   \fi
\ifx \showURL      \undefined \def \showURL       {\relax}        \fi
% invisible to TeX
\providecommand\bibfield[2]{#2}
\providecommand\bibinfo[2]{#2}
\providecommand\natexlab[1]{#1}
\providecommand\showeprint[2][]{arXiv:#2}

\bibitem[Alshaina et~al\mbox{.}(2017)]%
        {alshaina2017multi}
\bibfield{author}{\bibinfo{person}{S Alshaina}, \bibinfo{person}{Ansamma John},
  {and} \bibinfo{person}{Aneesh~G Nath}.} \bibinfo{year}{2017}\natexlab{}.
\newblock \showarticletitle{Multi-document abstractive summarization based on
  predicate argument structure}. In \bibinfo{booktitle}{\emph{2017 IEEE
  International Conference on Signal Processing, Informatics, Communication and
  Energy Systems (SPICES)}}. IEEE, \bibinfo{pages}{1--6}.
\newblock


\bibitem[Ammanabrolu et~al\mbox{.}(2021)]%
        {DBLP:conf/aaai/AmmanabroluCBR21}
\bibfield{author}{\bibinfo{person}{Prithviraj Ammanabrolu},
  \bibinfo{person}{Wesley Cheung}, \bibinfo{person}{William Broniec}, {and}
  \bibinfo{person}{Mark~O. Riedl}.} \bibinfo{year}{2021}\natexlab{}.
\newblock \showarticletitle{Automated Storytelling via Causal, Commonsense Plot
  Ordering}. In \bibinfo{booktitle}{\emph{Thirty-Fifth {AAAI} Conference on
  Artificial Intelligence, {AAAI} 2021, Thirty-Third Conference on Innovative
  Applications of Artificial Intelligence, {IAAI} 2021, The Eleventh Symposium
  on Educational Advances in Artificial Intelligence, {EAAI} 2021, Virtual
  Event, February 2-9, 2021}}. \bibinfo{pages}{5859--5867}.
\newblock
\urldef\tempurl%
\url{https://ojs.aaai.org/index.php/AAAI/article/view/16733}
\showURL{%
\tempurl}


\bibitem[Amplayo et~al\mbox{.}(2021)]%
        {DBLP:conf/aaai/AmplayoAL21}
\bibfield{author}{\bibinfo{person}{Reinald~Kim Amplayo},
  \bibinfo{person}{Stefanos Angelidis}, {and} \bibinfo{person}{Mirella
  Lapata}.} \bibinfo{year}{2021}\natexlab{}.
\newblock \showarticletitle{Unsupervised Opinion Summarization with Content
  Planning}. In \bibinfo{booktitle}{\emph{Thirty-Fifth {AAAI} Conference on
  Artificial Intelligence, {AAAI} 2021, Thirty-Third Conference on Innovative
  Applications of Artificial Intelligence, {IAAI} 2021, The Eleventh Symposium
  on Educational Advances in Artificial Intelligence, {EAAI} 2021, Virtual
  Event, February 2-9, 2021}}. \bibinfo{pages}{12489--12497}.
\newblock
\urldef\tempurl%
\url{https://ojs.aaai.org/index.php/AAAI/article/view/17481}
\showURL{%
\tempurl}


\bibitem[Azadani et~al\mbox{.}(2018)]%
        {DBLP:journals/jbi/AzadaniGD18}
\bibfield{author}{\bibinfo{person}{Mozhgan~Nasr Azadani},
  \bibinfo{person}{Nasser Ghadiri}, {and} \bibinfo{person}{Ensieh Davoodijam}.}
  \bibinfo{year}{2018}\natexlab{}.
\newblock \showarticletitle{Graph-based biomedical text summarization: An
  itemset mining and sentence clustering approach}.
\newblock \bibinfo{journal}{\emph{J. Biomed. Informatics}}
  \bibinfo{volume}{84} (\bibinfo{year}{2018}), \bibinfo{pages}{42--58}.
\newblock
\urldef\tempurl%
\url{https://doi.org/10.1016/j.jbi.2018.06.005}
\showURL{%
\tempurl}


\bibitem[Azaria et~al\mbox{.}(2016)]%
        {DBLP:conf/aaai/AzariaKM16}
\bibfield{author}{\bibinfo{person}{Amos Azaria}, \bibinfo{person}{Jayant
  Krishnamurthy}, {and} \bibinfo{person}{Tom~M. Mitchell}.}
  \bibinfo{year}{2016}\natexlab{}.
\newblock \showarticletitle{Instructable Intelligent Personal Agent}. In
  \bibinfo{booktitle}{\emph{Proceedings of the Thirtieth {AAAI} Conference on
  Artificial Intelligence, February 12-17, 2016, Phoenix, Arizona, {USA}}},
  \bibfield{editor}{\bibinfo{person}{Dale Schuurmans} {and}
  \bibinfo{person}{Michael~P. Wellman}} (Eds.). \bibinfo{pages}{2681--2689}.
\newblock
\urldef\tempurl%
\url{http://www.aaai.org/ocs/index.php/AAAI/AAAI16/paper/view/12383}
\showURL{%
\tempurl}


\bibitem[Azaria et~al\mbox{.}(2020)]%
        {DBLP:journals/aamas/AzariaSKLM20}
\bibfield{author}{\bibinfo{person}{Amos Azaria}, \bibinfo{person}{Shashank
  Srivastava}, \bibinfo{person}{Jayant Krishnamurthy}, \bibinfo{person}{Igor
  Labutov}, {and} \bibinfo{person}{Tom~M. Mitchell}.}
  \bibinfo{year}{2020}\natexlab{}.
\newblock \showarticletitle{An agent for learning new natural language
  commands}.
\newblock \bibinfo{journal}{\emph{Auton. Agents Multi Agent Syst.}}
  \bibinfo{volume}{34}, \bibinfo{number}{1} (\bibinfo{year}{2020}),
  \bibinfo{pages}{6}.
\newblock
\urldef\tempurl%
\url{https://doi.org/10.1007/s10458-019-09425-x}
\showDOI{\tempurl}


\bibitem[Bahdanau et~al\mbox{.}(2017)]%
        {DBLP:conf/iclr/BahdanauBXGLPCB17}
\bibfield{author}{\bibinfo{person}{Dzmitry Bahdanau}, \bibinfo{person}{Philemon
  Brakel}, \bibinfo{person}{Kelvin Xu}, \bibinfo{person}{Anirudh Goyal},
  \bibinfo{person}{Ryan Lowe}, \bibinfo{person}{Joelle Pineau},
  \bibinfo{person}{Aaron~C. Courville}, {and} \bibinfo{person}{Yoshua Bengio}.}
  \bibinfo{year}{2017}\natexlab{}.
\newblock \showarticletitle{An Actor-Critic Algorithm for Sequence Prediction}.
  In \bibinfo{booktitle}{\emph{5th International Conference on Learning
  Representations, {ICLR} 2017, Toulon, France, April 24-26, 2017, Conference
  Track Proceedings}}.
\newblock
\urldef\tempurl%
\url{https://openreview.net/forum?id=SJDaqqveg}
\showURL{%
\tempurl}


\bibitem[Banerjee et~al\mbox{.}(2015)]%
        {DBLP:conf/ijcai/BanerjeeMS15}
\bibfield{author}{\bibinfo{person}{Siddhartha Banerjee},
  \bibinfo{person}{Prasenjit Mitra}, {and} \bibinfo{person}{Kazunari
  Sugiyama}.} \bibinfo{year}{2015}\natexlab{}.
\newblock \showarticletitle{Multi-Document Abstractive Summarization Using
  {ILP} Based Multi-Sentence Compression}. In
  \bibinfo{booktitle}{\emph{Proceedings of the Twenty-Fourth International
  Joint Conference on Artificial Intelligence, {IJCAI} 2015, Buenos Aires,
  Argentina, July 25-31, 2015}}, \bibfield{editor}{\bibinfo{person}{Qiang Yang}
  {and} \bibinfo{person}{Michael~J. Wooldridge}} (Eds.).
  \bibinfo{pages}{1208--1214}.
\newblock
\urldef\tempurl%
\url{http://ijcai.org/Abstract/15/174}
\showURL{%
\tempurl}


\bibitem[Baud et~al\mbox{.}(1993)]%
        {baud1993modelling}
\bibfield{author}{\bibinfo{person}{Robert Baud}, \bibinfo{person}{Christian
  Lovis}, \bibinfo{person}{Laurence Alpay}, \bibinfo{person}{Anne-Marie
  Rassinoux}, \bibinfo{person}{JR Scherrer}, \bibinfo{person}{Anthony Nowlan},
  {and} \bibinfo{person}{Alan Rector}.} \bibinfo{year}{1993}\natexlab{}.
\newblock \showarticletitle{Modelling for natural language understanding.}. In
  \bibinfo{booktitle}{\emph{Proceedings of the Annual Symposium on Computer
  Application in Medical Care}}. American Medical Informatics Association,
  \bibinfo{pages}{289}.
\newblock


\bibitem[Bisk et~al\mbox{.}(2018)]%
        {DBLP:conf/aaai/BiskSCM18}
\bibfield{author}{\bibinfo{person}{Yonatan Bisk}, \bibinfo{person}{Kevin~J.
  Shih}, \bibinfo{person}{Yejin Choi}, {and} \bibinfo{person}{Daniel Marcu}.}
  \bibinfo{year}{2018}\natexlab{}.
\newblock \showarticletitle{Learning Interpretable Spatial Operations in a Rich
  3D Blocks World}. In \bibinfo{booktitle}{\emph{Proceedings of the
  Thirty-Second {AAAI} Conference on Artificial Intelligence, (AAAI-18), the
  30th innovative Applications of Artificial Intelligence (IAAI-18), and the
  8th {AAAI} Symposium on Educational Advances in Artificial Intelligence
  (EAAI-18), New Orleans, Louisiana, USA, February 2-7, 2018}},
  \bibfield{editor}{\bibinfo{person}{Sheila~A. McIlraith} {and}
  \bibinfo{person}{Kilian~Q. Weinberger}} (Eds.). \bibinfo{pages}{5028--5036}.
\newblock
\urldef\tempurl%
\url{https://www.aaai.org/ocs/index.php/AAAI/AAAI18/paper/view/17410}
\showURL{%
\tempurl}


\bibitem[Blindheim et~al\mbox{.}(2020)]%
        {blindheim2020risk}
\bibfield{author}{\bibinfo{person}{Simon Blindheim}, \bibinfo{person}{Sebastien
  Gros}, {and} \bibinfo{person}{Tor~Arne Johansen}.}
  \bibinfo{year}{2020}\natexlab{}.
\newblock \showarticletitle{Risk-based model predictive control for autonomous
  ship emergency management}.
\newblock \bibinfo{journal}{\emph{IFAC-PapersOnLine}} \bibinfo{volume}{53},
  \bibinfo{number}{2} (\bibinfo{year}{2020}), \bibinfo{pages}{14524--14531}.
\newblock


\bibitem[Blukis et~al\mbox{.}(2018)]%
        {DBLP:conf/corl/BlukisMKA18}
\bibfield{author}{\bibinfo{person}{Valts Blukis},
  \bibinfo{person}{Dipendra~Kumar Misra}, \bibinfo{person}{Ross~A. Knepper},
  {and} \bibinfo{person}{Yoav Artzi}.} \bibinfo{year}{2018}\natexlab{}.
\newblock \showarticletitle{Mapping Navigation Instructions to Continuous
  Control Actions with Position-Visitation Prediction}. In
  \bibinfo{booktitle}{\emph{2nd Annual Conference on Robot Learning, CoRL 2018,
  Z{\"{u}}rich, Switzerland, 29-31 October 2018, Proceedings}}
  \emph{(\bibinfo{series}{Proceedings of Machine Learning Research},
  Vol.~\bibinfo{volume}{87})}. \bibinfo{pages}{505--518}.
\newblock
\urldef\tempurl%
\url{http://proceedings.mlr.press/v87/blukis18a.html}
\showURL{%
\tempurl}


\bibitem[Branavan et~al\mbox{.}(2009)]%
        {DBLP:conf/acl/BranavanCZB09}
\bibfield{author}{\bibinfo{person}{S.~R.~K. Branavan}, \bibinfo{person}{Harr
  Chen}, \bibinfo{person}{Luke~S. Zettlemoyer}, {and} \bibinfo{person}{Regina
  Barzilay}.} \bibinfo{year}{2009}\natexlab{}.
\newblock \showarticletitle{Reinforcement Learning for Mapping Instructions to
  Actions}. In \bibinfo{booktitle}{\emph{{ACL} 2009, Proceedings of the 47th
  Annual Meeting of the Association for Computational Linguistics and the 4th
  International Joint Conference on Natural Language Processing of the AFNLP,
  2-7 August 2009, Singapore}}, \bibfield{editor}{\bibinfo{person}{Keh{-}Yih
  Su}, \bibinfo{person}{Jian Su}, {and} \bibinfo{person}{Janyce Wiebe}} (Eds.).
  \bibinfo{pages}{82--90}.
\newblock
\urldef\tempurl%
\url{https://aclanthology.org/P09-1010/}
\showURL{%
\tempurl}


\bibitem[Branavan et~al\mbox{.}(2012)]%
        {DBLP:conf/acl/BranavanKLB12}
\bibfield{author}{\bibinfo{person}{S.~R.~K. Branavan}, \bibinfo{person}{Nate
  Kushman}, \bibinfo{person}{Tao Lei}, {and} \bibinfo{person}{Regina
  Barzilay}.} \bibinfo{year}{2012}\natexlab{}.
\newblock \showarticletitle{Learning High-Level Planning from Text}. In
  \bibinfo{booktitle}{\emph{The 50th Annual Meeting of the Association for
  Computational Linguistics, Proceedings of the Conference, July 8-14, 2012,
  Jeju Island, Korea - Volume 1: Long Papers}}. \bibinfo{pages}{126--135}.
\newblock
\urldef\tempurl%
\url{https://aclanthology.org/P12-1014/}
\showURL{%
\tempurl}


\bibitem[Brawer et~al\mbox{.}(2018)]%
        {DBLP:conf/iros/BrawerMRWS18}
\bibfield{author}{\bibinfo{person}{Jake Brawer}, \bibinfo{person}{Olivier
  Mangin}, \bibinfo{person}{Alessandro Roncone}, \bibinfo{person}{Sarah
  Widder}, {and} \bibinfo{person}{Brian Scassellati}.}
  \bibinfo{year}{2018}\natexlab{}.
\newblock \showarticletitle{Situated Human-Robot Collaboration: predicting
  intent from grounded natural language}. In \bibinfo{booktitle}{\emph{2018
  {IEEE/RSJ} International Conference on Intelligent Robots and Systems, {IROS}
  2018, Madrid, Spain, October 1-5, 2018}}. \bibinfo{pages}{827--833}.
\newblock
\urldef\tempurl%
\url{https://doi.org/10.1109/IROS.2018.8593942}
\showURL{%
\tempurl}


\bibitem[Buzhinsky(2019)]%
        {DBLP:conf/indin/Buzhinsky19}
\bibfield{author}{\bibinfo{person}{Igor Buzhinsky}.}
  \bibinfo{year}{2019}\natexlab{}.
\newblock \showarticletitle{Formalization of natural language requirements into
  temporal logics: a survey}. In \bibinfo{booktitle}{\emph{17th {IEEE}
  International Conference on Industrial Informatics, {INDIN} 2019, Helsinki,
  Finland, July 22-25, 2019}}. \bibinfo{pages}{400--406}.
\newblock
\urldef\tempurl%
\url{https://doi.org/10.1109/INDIN41052.2019.8972130}
\showURL{%
\tempurl}


\bibitem[Cambria and White(2014)]%
        {cambria2014jumping}
\bibfield{author}{\bibinfo{person}{Erik Cambria} {and} \bibinfo{person}{Bebo
  White}.} \bibinfo{year}{2014}\natexlab{}.
\newblock \showarticletitle{Jumping NLP curves: A review of natural language
  processing research}.
\newblock \bibinfo{journal}{\emph{IEEE Computational intelligence magazine}}
  \bibinfo{volume}{9}, \bibinfo{number}{2} (\bibinfo{year}{2014}),
  \bibinfo{pages}{48--57}.
\newblock


\bibitem[Camilleri(2002)]%
        {camilleri2002dialogue}
\bibfield{author}{\bibinfo{person}{Guy Camilleri}.}
  \bibinfo{year}{2002}\natexlab{}.
\newblock \showarticletitle{Dialogue systems and planning}. In
  \bibinfo{booktitle}{\emph{International Conference on Text, Speech and
  Dialogue}}. Springer, \bibinfo{pages}{429--436}.
\newblock


\bibitem[Cantrell et~al\mbox{.}(2011)]%
        {DBLP:conf/ro-man/CantrellSS11}
\bibfield{author}{\bibinfo{person}{Rehj Cantrell}, \bibinfo{person}{Paul~W.
  Schermerhorn}, {and} \bibinfo{person}{Matthias Scheutz}.}
  \bibinfo{year}{2011}\natexlab{}.
\newblock \showarticletitle{Learning actions from human-robot dialogues}. In
  \bibinfo{booktitle}{\emph{20th {IEEE} International Symposium on Robot and
  Human Interactive Communication, {RO-MAN} 2011, Atlanta, Georgia, USA, July
  31 - August 3, 2011}}, \bibfield{editor}{\bibinfo{person}{Henrik~I.
  Christensen}} (Ed.). \bibinfo{pages}{125--130}.
\newblock
\urldef\tempurl%
\url{https://doi.org/10.1109/ROMAN.2011.6005199}
\showURL{%
\tempurl}


\bibitem[Cantrell et~al\mbox{.}(2012)]%
        {DBLP:conf/hri/CantrellTSBKS12}
\bibfield{author}{\bibinfo{person}{Rehj Cantrell}, \bibinfo{person}{Kartik
  Talamadupula}, \bibinfo{person}{Paul~W. Schermerhorn}, \bibinfo{person}{J.
  Benton}, \bibinfo{person}{Subbarao Kambhampati}, {and}
  \bibinfo{person}{Matthias Scheutz}.} \bibinfo{year}{2012}\natexlab{}.
\newblock \showarticletitle{Tell me when and why to do it!: run-time planner
  model updates via natural language instruction}. In
  \bibinfo{booktitle}{\emph{HRI}}, \bibfield{editor}{\bibinfo{person}{Holly~A.
  Yanco}, \bibinfo{person}{Aaron Steinfeld}, \bibinfo{person}{Vanessa Evers},
  {and} \bibinfo{person}{Odest~Chadwicke Jenkins}} (Eds.).
  \bibinfo{pages}{471--478}.
\newblock
\urldef\tempurl%
\url{https://doi.org/10.1145/2157689.2157840}
\showURL{%
\tempurl}


\bibitem[Cercone and McCalla(1986)]%
        {cercone1986accessing}
\bibfield{author}{\bibinfo{person}{Nick Cercone} {and} \bibinfo{person}{Gordon
  McCalla}.} \bibinfo{year}{1986}\natexlab{}.
\newblock \showarticletitle{Accessing knowledge through natural language}.
\newblock In \bibinfo{booktitle}{\emph{Advances in Computers}}.
  Vol.~\bibinfo{volume}{25}. \bibinfo{publisher}{Elsevier},
  \bibinfo{pages}{1--99}.
\newblock


\bibitem[Chen and Mooney(2011)]%
        {DBLP:conf/aaai/ChenM11}
\bibfield{author}{\bibinfo{person}{David~L. Chen} {and}
  \bibinfo{person}{Raymond~J. Mooney}.} \bibinfo{year}{2011}\natexlab{}.
\newblock \showarticletitle{Learning to Interpret Natural Language Navigation
  Instructions from Observations}. In \bibinfo{booktitle}{\emph{Proceedings of
  the Twenty-Fifth {AAAI} Conference on Artificial Intelligence, {AAAI} 2011,
  San Francisco, California, USA, August 7-11, 2011}},
  \bibfield{editor}{\bibinfo{person}{Wolfram Burgard} {and}
  \bibinfo{person}{Dan Roth}} (Eds.).
\newblock
\urldef\tempurl%
\url{http://www.aaai.org/ocs/index.php/AAAI/AAAI11/paper/view/3701}
\showURL{%
\tempurl}


\bibitem[Chen et~al\mbox{.}(2017)]%
        {DBLP:journals/sigkdd/ChenLYT17}
\bibfield{author}{\bibinfo{person}{Hongshen Chen}, \bibinfo{person}{Xiaorui
  Liu}, \bibinfo{person}{Dawei Yin}, {and} \bibinfo{person}{Jiliang Tang}.}
  \bibinfo{year}{2017}\natexlab{}.
\newblock \showarticletitle{A Survey on Dialogue Systems: Recent Advances and
  New Frontiers}.
\newblock \bibinfo{journal}{\emph{{SIGKDD} Explor.}} \bibinfo{volume}{19},
  \bibinfo{number}{2} (\bibinfo{year}{2017}), \bibinfo{pages}{25--35}.
\newblock
\urldef\tempurl%
\url{https://doi.org/10.1145/3166054.3166058}
\showURL{%
\tempurl}


\bibitem[Chen et~al\mbox{.}(2019)]%
        {DBLP:conf/cvpr/ChenSMSA19}
\bibfield{author}{\bibinfo{person}{Howard Chen}, \bibinfo{person}{Alane Suhr},
  \bibinfo{person}{Dipendra Misra}, \bibinfo{person}{Noah Snavely}, {and}
  \bibinfo{person}{Yoav Artzi}.} \bibinfo{year}{2019}\natexlab{}.
\newblock \showarticletitle{{TOUCHDOWN:} Natural Language Navigation and
  Spatial Reasoning in Visual Street Environments}. In
  \bibinfo{booktitle}{\emph{{IEEE} Conference on Computer Vision and Pattern
  Recognition, {CVPR} 2019, Long Beach, CA, USA, June 16-20, 2019}}.
  \bibinfo{pages}{12538--12547}.
\newblock
\urldef\tempurl%
\url{http://openaccess.thecvf.com/content\_CVPR\_2019/html/Chen\_TOUCHDOWN\_Natural\_Language\_Navigation\_and\_Spatial\_Reasoning\_in\_Visual\_Street\_CVPR\_2019\_paper.html}
\showURL{%
\tempurl}


\bibitem[Chen et~al\mbox{.}(2020)]%
        {DBLP:conf/icra/ChenTKBA20}
\bibfield{author}{\bibinfo{person}{Haonan Chen}, \bibinfo{person}{Hao Tan},
  \bibinfo{person}{Alan Kuntz}, \bibinfo{person}{Mohit Bansal}, {and}
  \bibinfo{person}{Ron Alterovitz}.} \bibinfo{year}{2020}\natexlab{}.
\newblock \showarticletitle{Enabling Robots to Understand Incomplete Natural
  Language Instructions Using Commonsense Reasoning}. In
  \bibinfo{booktitle}{\emph{2020 {IEEE} International Conference on Robotics
  and Automation, {ICRA} 2020, Paris, France, May 31 - August 31, 2020}}.
  \bibinfo{pages}{1963--1969}.
\newblock
\urldef\tempurl%
\url{https://doi.org/10.1109/ICRA40945.2020.9197315}
\showURL{%
\tempurl}


\bibitem[Chi et~al\mbox{.}(2020)]%
        {DBLP:conf/aaai/ChiSEKH20}
\bibfield{author}{\bibinfo{person}{Ta{-}Chung Chi}, \bibinfo{person}{Minmin
  Shen}, \bibinfo{person}{Mihail Eric}, \bibinfo{person}{Seokhwan Kim}, {and}
  \bibinfo{person}{Dilek Hakkani{-}T{\"{u}}r}.}
  \bibinfo{year}{2020}\natexlab{}.
\newblock \showarticletitle{Just Ask: An Interactive Learning Framework for
  Vision and Language Navigation}. In \bibinfo{booktitle}{\emph{The
  Thirty-Fourth {AAAI} Conference on Artificial Intelligence, {AAAI} 2020, The
  Thirty-Second Innovative Applications of Artificial Intelligence Conference,
  {IAAI} 2020, The Tenth {AAAI} Symposium on Educational Advances in Artificial
  Intelligence, {EAAI} 2020, New York, NY, USA, February 7-12, 2020}}.
  \bibinfo{pages}{2459--2466}.
\newblock
\urldef\tempurl%
\url{https://ojs.aaai.org/index.php/AAAI/article/view/5627}
\showURL{%
\tempurl}


\bibitem[Chu-Carroll and Carberry(1994)]%
        {chu1994plan}
\bibfield{author}{\bibinfo{person}{Jennifer Chu-Carroll} {and}
  \bibinfo{person}{Sandra Carberry}.} \bibinfo{year}{1994}\natexlab{}.
\newblock \showarticletitle{A plan-based model for response generation in
  collaborative task-oriented dialogues}.
\newblock \bibinfo{journal}{\emph{arXiv preprint cmp-lg/9405011}}
  (\bibinfo{year}{1994}).
\newblock


\bibitem[Cohen(2020)]%
        {cohen2020back}
\bibfield{author}{\bibinfo{person}{Philip~R Cohen}.}
  \bibinfo{year}{2020}\natexlab{}.
\newblock \showarticletitle{Back to the future for dialogue research}. In
  \bibinfo{booktitle}{\emph{Proceedings of the AAAI Conference on Artificial
  Intelligence}}, Vol.~\bibinfo{volume}{34}. \bibinfo{pages}{13514--13519}.
\newblock


\bibitem[Costa et~al\mbox{.}(2022)]%
        {costa2022survey}
\bibfield{author}{\bibinfo{person}{Daniel~G Costa}, \bibinfo{person}{Jo{\~a}o
  Paulo~J Peixoto}, \bibinfo{person}{Thiago~C Jesus}, \bibinfo{person}{Paulo
  Portugal}, \bibinfo{person}{Francisco Vasques}, \bibinfo{person}{Elivelton
  Rangel}, {and} \bibinfo{person}{Maycon Peixoto}.}
  \bibinfo{year}{2022}\natexlab{}.
\newblock \showarticletitle{A Survey of Emergencies Management Systems in Smart
  Cities}.
\newblock \bibinfo{journal}{\emph{IEEE Access}} (\bibinfo{year}{2022}).
\newblock


\bibitem[Dong et~al\mbox{.}(2021)]%
        {dong2021survey}
\bibfield{author}{\bibinfo{person}{Chenhe Dong}, \bibinfo{person}{Yinghui Li},
  \bibinfo{person}{Haifan Gong}, \bibinfo{person}{Miaoxin Chen},
  \bibinfo{person}{Junxin Li}, \bibinfo{person}{Ying Shen}, {and}
  \bibinfo{person}{Min Yang}.} \bibinfo{year}{2021}\natexlab{}.
\newblock \showarticletitle{A Survey of Natural Language Generation}.
\newblock \bibinfo{journal}{\emph{arXiv preprint arXiv:2112.11739}}
  (\bibinfo{year}{2021}).
\newblock


\bibitem[Fan et~al\mbox{.}(2019)]%
        {DBLP:conf/acl/FanLD19}
\bibfield{author}{\bibinfo{person}{Angela Fan}, \bibinfo{person}{Mike Lewis},
  {and} \bibinfo{person}{Yann~N. Dauphin}.} \bibinfo{year}{2019}\natexlab{}.
\newblock \showarticletitle{Strategies for Structuring Story Generation}. In
  \bibinfo{booktitle}{\emph{Proceedings of the 57th Conference of the
  Association for Computational Linguistics, {ACL} 2019, Florence, Italy, July
  28- August 2, 2019, Volume 1: Long Papers}},
  \bibfield{editor}{\bibinfo{person}{Anna Korhonen}, \bibinfo{person}{David~R.
  Traum}, {and} \bibinfo{person}{Llu{\'{\i}}s M{\`{a}}rquez}} (Eds.).
  \bibinfo{pages}{2650--2660}.
\newblock
\urldef\tempurl%
\url{https://doi.org/10.18653/v1/p19-1254}
\showURL{%
\tempurl}


\bibitem[Feng et~al\mbox{.}(2018)]%
        {DBLP:conf/ijcai/FengZK18}
\bibfield{author}{\bibinfo{person}{Wenfeng Feng}, \bibinfo{person}{Hankz~Hankui
  Zhuo}, {and} \bibinfo{person}{Subbarao Kambhampati}.}
  \bibinfo{year}{2018}\natexlab{}.
\newblock \showarticletitle{Extracting Action Sequences from Texts Based on
  Deep Reinforcement Learning}. In \bibinfo{booktitle}{\emph{{IJCAI}}}.
  \bibinfo{pages}{4064--4070}.
\newblock


\bibitem[Fogli and Guida(2013)]%
        {fogli2013knowledge}
\bibfield{author}{\bibinfo{person}{Daniela Fogli} {and}
  \bibinfo{person}{Giovanni Guida}.} \bibinfo{year}{2013}\natexlab{}.
\newblock \showarticletitle{Knowledge-centered design of decision support
  systems for emergency management}.
\newblock \bibinfo{journal}{\emph{Decision Support Systems}}
  \bibinfo{volume}{55}, \bibinfo{number}{1} (\bibinfo{year}{2013}),
  \bibinfo{pages}{336--347}.
\newblock


\bibitem[Fox and Long(2002)]%
        {pddl+}
\bibfield{author}{\bibinfo{person}{Maria Fox} {and} \bibinfo{person}{Derek
  Long}.} \bibinfo{year}{2002}\natexlab{}.
\newblock \showarticletitle{PDDL+: Modeling continuous time dependent effects}.
  In \bibinfo{booktitle}{\emph{Proceedings of the 3rd International NASA
  Workshop on Planning and Scheduling for Space}}, Vol.~\bibinfo{volume}{4}.
  \bibinfo{pages}{34}.
\newblock


\bibitem[Frasca et~al\mbox{.}(2021)]%
        {DBLP:conf/aaai/FrascaOCS21}
\bibfield{author}{\bibinfo{person}{Tyler~M. Frasca}, \bibinfo{person}{Bradley
  Oosterveld}, \bibinfo{person}{Meia Chita{-}Tegmark}, {and}
  \bibinfo{person}{Matthias Scheutz}.} \bibinfo{year}{2021}\natexlab{}.
\newblock \showarticletitle{Enabling Fast Instruction-Based Modification of
  Learned Robot Skills}. In \bibinfo{booktitle}{\emph{Thirty-Fifth {AAAI}
  Conference on Artificial Intelligence, {AAAI} 2021, Thirty-Third Conference
  on Innovative Applications of Artificial Intelligence, {IAAI} 2021, The
  Eleventh Symposium on Educational Advances in Artificial Intelligence, {EAAI}
  2021, Virtual Event, February 2-9, 2021}}. \bibinfo{pages}{6075--6083}.
\newblock
\urldef\tempurl%
\url{https://ojs.aaai.org/index.php/AAAI/article/view/16757}
\showURL{%
\tempurl}


\bibitem[Garoufi(2014)]%
        {DBLP:journals/llc/Garoufi14}
\bibfield{author}{\bibinfo{person}{Konstantina Garoufi}.}
  \bibinfo{year}{2014}\natexlab{}.
\newblock \showarticletitle{Planning-Based Models of Natural Language
  Generation}.
\newblock \bibinfo{journal}{\emph{Lang. Linguistics Compass}}
  \bibinfo{volume}{8}, \bibinfo{number}{1} (\bibinfo{year}{2014}),
  \bibinfo{pages}{1--10}.
\newblock
\urldef\tempurl%
\url{https://doi.org/10.1111/lnc3.12053}
\showURL{%
\tempurl}


\bibitem[Gatt and Krahmer(2018)]%
        {gatt2018survey}
\bibfield{author}{\bibinfo{person}{Albert Gatt} {and} \bibinfo{person}{Emiel
  Krahmer}.} \bibinfo{year}{2018}\natexlab{}.
\newblock \showarticletitle{Survey of the state of the art in natural language
  generation: Core tasks, applications and evaluation}.
\newblock \bibinfo{journal}{\emph{Journal of Artificial Intelligence Research}}
   \bibinfo{volume}{61} (\bibinfo{year}{2018}), \bibinfo{pages}{65--170}.
\newblock


\bibitem[Geib and Steedman(2007)]%
        {DBLP:conf/ijcai/GeibS07}
\bibfield{author}{\bibinfo{person}{Christopher~W. Geib} {and}
  \bibinfo{person}{Mark Steedman}.} \bibinfo{year}{2007}\natexlab{}.
\newblock \showarticletitle{On Natural Language Processing and Plan
  Recognition}. In \bibinfo{booktitle}{\emph{{IJCAI} 2007, Proceedings of the
  20th International Joint Conference on Artificial Intelligence, Hyderabad,
  India, January 6-12, 2007}}, \bibfield{editor}{\bibinfo{person}{Manuela~M.
  Veloso}} (Ed.). \bibinfo{pages}{1612--1617}.
\newblock
\urldef\tempurl%
\url{http://ijcai.org/Proceedings/07/Papers/260.pdf}
\showURL{%
\tempurl}


\bibitem[Gerevini and Long(2005)]%
        {pddl3}
\bibfield{author}{\bibinfo{person}{Alfonso Gerevini} {and}
  \bibinfo{person}{Derek Long}.} \bibinfo{year}{2005}\natexlab{}.
\newblock \bibinfo{booktitle}{\emph{Plan constraints and preferences in
  PDDL3}}.
\newblock \bibinfo{type}{{T}echnical {R}eport}. \bibinfo{institution}{Technical
  Report 2005-08-07, Department of Electronics for Automation~…}.
\newblock


\bibitem[Guan et~al\mbox{.}(2021)]%
        {guan2021widening}
\bibfield{author}{\bibinfo{person}{Lin Guan}, \bibinfo{person}{Mudit Verma},
  \bibinfo{person}{Sihang Guo}, \bibinfo{person}{Ruohan Zhang}, {and}
  \bibinfo{person}{Subbarao Kambhampati}.} \bibinfo{year}{2021}\natexlab{}.
\newblock \showarticletitle{Widening the Pipeline in Human-Guided Reinforcement
  Learning with Explanation and Context-Aware Data Augmentation}.
\newblock \bibinfo{journal}{\emph{Advances in Neural Information Processing
  Systems}}  \bibinfo{volume}{34} (\bibinfo{year}{2021}).
\newblock


\bibitem[G{\"{u}}l{\c{c}}ehre et~al\mbox{.}(2017)]%
        {DBLP:conf/rep4nlp/GulcehreDTB17}
\bibfield{author}{\bibinfo{person}{{\c{C}}aglar G{\"{u}}l{\c{c}}ehre},
  \bibinfo{person}{Francis Dutil}, \bibinfo{person}{Adam Trischler}, {and}
  \bibinfo{person}{Yoshua Bengio}.} \bibinfo{year}{2017}\natexlab{}.
\newblock \showarticletitle{Plan, Attend, Generate: Character-Level Neural
  Machine Translation with Planning}. In \bibinfo{booktitle}{\emph{Proceedings
  of the 2nd Workshop on Representation Learning for NLP, Rep4NLP@ACL 2017,
  Vancouver, Canada, August 3, 2017}}, \bibfield{editor}{\bibinfo{person}{Phil
  Blunsom}, \bibinfo{person}{Antoine Bordes}, \bibinfo{person}{Kyunghyun Cho},
  \bibinfo{person}{Shay~B. Cohen}, \bibinfo{person}{Chris Dyer},
  \bibinfo{person}{Edward Grefenstette}, \bibinfo{person}{Karl~Moritz Hermann},
  \bibinfo{person}{Laura Rimell}, \bibinfo{person}{Jason Weston}, {and}
  \bibinfo{person}{Scott Yih}} (Eds.). \bibinfo{pages}{228--234}.
\newblock
\urldef\tempurl%
\url{https://doi.org/10.18653/v1/w17-2627}
\showURL{%
\tempurl}


\bibitem[Hayton et~al\mbox{.}(2020)]%
        {DBLP:conf/aaai/HaytonPFL20}
\bibfield{author}{\bibinfo{person}{Thomas Hayton}, \bibinfo{person}{Julie
  Porteous}, \bibinfo{person}{Jo{\~{a}}o~Fernando Ferreira}, {and}
  \bibinfo{person}{Alan Lindsay}.} \bibinfo{year}{2020}\natexlab{}.
\newblock \showarticletitle{Narrative Planning Model Acquisition from Text
  Summaries and Descriptions}. In \bibinfo{booktitle}{\emph{{AAAI}}}.
  \bibinfo{pages}{1709--1716}.
\newblock


\bibitem[Hl{\'a}dek et~al\mbox{.}(2020)]%
        {hladek2020survey}
\bibfield{author}{\bibinfo{person}{Daniel Hl{\'a}dek}, \bibinfo{person}{J{\'a}n
  Sta{\v{s}}}, {and} \bibinfo{person}{Mat{\'u}{\v{s}} Pleva}.}
  \bibinfo{year}{2020}\natexlab{}.
\newblock \showarticletitle{Survey of automatic spelling correction}.
\newblock \bibinfo{journal}{\emph{Electronics}} \bibinfo{volume}{9},
  \bibinfo{number}{10} (\bibinfo{year}{2020}), \bibinfo{pages}{1670}.
\newblock


\bibitem[Hua et~al\mbox{.}(2021)]%
        {hua-etal-2021-dyploc}
\bibfield{author}{\bibinfo{person}{Xinyu Hua}, \bibinfo{person}{Ashwin
  Sreevatsa}, {and} \bibinfo{person}{Lu Wang}.}
  \bibinfo{year}{2021}\natexlab{}.
\newblock \showarticletitle{{DYPLOC}: Dynamic Planning of Content Using Mixed
  Language Models for Text Generation}. In
  \bibinfo{booktitle}{\emph{Proceedings of {ACL}}}.
  \bibinfo{pages}{6408--6423}.
\newblock


\bibitem[Huang et~al\mbox{.}(2020)]%
        {DBLP:conf/kdd/HuangWFZSL20}
\bibfield{author}{\bibinfo{person}{Jizhou Huang}, \bibinfo{person}{Haifeng
  Wang}, \bibinfo{person}{Miao Fan}, \bibinfo{person}{An Zhuo},
  \bibinfo{person}{Yibo Sun}, {and} \bibinfo{person}{Ying Li}.}
  \bibinfo{year}{2020}\natexlab{}.
\newblock \showarticletitle{Understanding the Impact of the {COVID-19} Pandemic
  on Transportation-related Behaviors with Human Mobility Data}.
\newblock  (\bibinfo{year}{2020}), \bibinfo{pages}{3443--3450}.
\newblock
\urldef\tempurl%
\url{https://doi.org/10.1145/3394486.3412856}
\showURL{%
\tempurl}


\bibitem[Huang et~al\mbox{.}(2022a)]%
        {DBLP:conf/icml/HuangAPM22}
\bibfield{author}{\bibinfo{person}{Wenlong Huang}, \bibinfo{person}{Pieter
  Abbeel}, \bibinfo{person}{Deepak Pathak}, {and} \bibinfo{person}{Igor
  Mordatch}.} \bibinfo{year}{2022}\natexlab{a}.
\newblock \showarticletitle{Language Models as Zero-Shot Planners: Extracting
  Actionable Knowledge for Embodied Agents}.
\newblock   \bibinfo{volume}{162} (\bibinfo{year}{2022}),
  \bibinfo{pages}{9118--9147}.
\newblock
\urldef\tempurl%
\url{https://proceedings.mlr.press/v162/huang22a.html}
\showURL{%
\tempurl}


\bibitem[Huang et~al\mbox{.}(2022b)]%
        {DBLP:journals/corr/abs-2207-05608}
\bibfield{author}{\bibinfo{person}{Wenlong Huang}, \bibinfo{person}{Fei Xia},
  \bibinfo{person}{Ted Xiao}, \bibinfo{person}{Harris Chan},
  \bibinfo{person}{Jacky Liang}, \bibinfo{person}{Pete Florence},
  \bibinfo{person}{Andy Zeng}, \bibinfo{person}{Jonathan Tompson},
  \bibinfo{person}{Igor Mordatch}, \bibinfo{person}{Yevgen Chebotar},
  \bibinfo{person}{Pierre Sermanet}, \bibinfo{person}{Noah Brown},
  \bibinfo{person}{Tomas Jackson}, \bibinfo{person}{Linda Luu},
  \bibinfo{person}{Sergey Levine}, \bibinfo{person}{Karol Hausman}, {and}
  \bibinfo{person}{Brian Ichter}.} \bibinfo{year}{2022}\natexlab{b}.
\newblock \showarticletitle{Inner Monologue: Embodied Reasoning through
  Planning with Language Models}.
\newblock \bibinfo{journal}{\emph{CoRR}}  \bibinfo{volume}{abs/2207.05608}
  (\bibinfo{year}{2022}).
\newblock
\urldef\tempurl%
\url{https://doi.org/10.48550/arXiv.2207.05608}
\showDOI{\tempurl}
\showeprint[arXiv]{2207.05608}


\bibitem[Hung and Yoshimi(2016)]%
        {hung2016extracting}
\bibfield{author}{\bibinfo{person}{Pham~Ngoc Hung} {and}
  \bibinfo{person}{Takashi Yoshimi}.} \bibinfo{year}{2016}\natexlab{}.
\newblock \showarticletitle{Extracting actions from instruction manual and
  testing their execution in a robotic simulation}.
\newblock \bibinfo{journal}{\emph{ASEAN Engineering Journal}}
  \bibinfo{volume}{6}, \bibinfo{number}{1} (\bibinfo{year}{2016}),
  \bibinfo{pages}{47--58}.
\newblock


\bibitem[Iqbal and Qureshi(2020)]%
        {iqbal2020survey}
\bibfield{author}{\bibinfo{person}{Touseef Iqbal} {and} \bibinfo{person}{Shaima
  Qureshi}.} \bibinfo{year}{2020}\natexlab{}.
\newblock \showarticletitle{The survey: Text generation models in deep
  learning}.
\newblock \bibinfo{journal}{\emph{Journal of King Saud University-Computer and
  Information Sciences}} (\bibinfo{year}{2020}).
\newblock


\bibitem[Jansen(2020)]%
        {DBLP:conf/emnlp/Jansen20}
\bibfield{author}{\bibinfo{person}{Peter~A. Jansen}.}
  \bibinfo{year}{2020}\natexlab{}.
\newblock \showarticletitle{Visually-Grounded Planning without Vision: Language
  Models Infer Detailed Plans from High-level Instructions}.
\newblock   \bibinfo{volume}{{EMNLP} 2020} (\bibinfo{year}{2020}),
  \bibinfo{pages}{4412--4417}.
\newblock
\urldef\tempurl%
\url{https://doi.org/10.18653/v1/2020.findings-emnlp.395}
\showDOI{\tempurl}


\bibitem[Jin et~al\mbox{.}(2020)]%
        {jin2020recent}
\bibfield{author}{\bibinfo{person}{HanQi Jin}, \bibinfo{person}{Yue Cao},
  \bibinfo{person}{TianMing Wang}, \bibinfo{person}{XinYu Xing}, {and}
  \bibinfo{person}{XiaoJun Wan}.} \bibinfo{year}{2020}\natexlab{}.
\newblock \showarticletitle{Recent advances of neural text generation: Core
  tasks, datasets, models and challenges}.
\newblock \bibinfo{journal}{\emph{Science China Technological Sciences}}
  \bibinfo{volume}{63}, \bibinfo{number}{10} (\bibinfo{year}{2020}),
  \bibinfo{pages}{1990--2010}.
\newblock


\bibitem[Kambhampati(2021)]%
        {DBLP:journals/cacm/Kambhampati21}
\bibfield{author}{\bibinfo{person}{Subbarao Kambhampati}.}
  \bibinfo{year}{2021}\natexlab{}.
\newblock \showarticletitle{Polanyi's revenge and AI's new romance with tacit
  knowledge}.
\newblock \bibinfo{journal}{\emph{Commun. {ACM}}} \bibinfo{volume}{64},
  \bibinfo{number}{2} (\bibinfo{year}{2021}), \bibinfo{pages}{31--32}.
\newblock
\urldef\tempurl%
\url{https://doi.org/10.1145/3446369}
\showURL{%
\tempurl}


\bibitem[Kambhampati et~al\mbox{.}(2021)]%
        {DBLP:journals/corr/abs-2109-09904}
\bibfield{author}{\bibinfo{person}{Subbarao Kambhampati},
  \bibinfo{person}{Sarath Sreedharan}, \bibinfo{person}{Mudit Verma},
  \bibinfo{person}{Yantian Zha}, {and} \bibinfo{person}{Lin Guan}.}
  \bibinfo{year}{2021}\natexlab{}.
\newblock \showarticletitle{Symbols as a Lingua Franca for Bridging Human-AI
  Chasm for Explainable and Advisable {AI} Systems}.
\newblock \bibinfo{journal}{\emph{CoRR}}  \bibinfo{volume}{abs/2109.09904}
  (\bibinfo{year}{2021}).
\newblock
\showeprint[arXiv]{2109.09904}
\urldef\tempurl%
\url{https://arxiv.org/abs/2109.09904}
\showURL{%
\tempurl}


\bibitem[Khurana et~al\mbox{.}(2022)]%
        {khurana2022natural}
\bibfield{author}{\bibinfo{person}{Diksha Khurana}, \bibinfo{person}{Aditya
  Koli}, \bibinfo{person}{Kiran Khatter}, {and} \bibinfo{person}{Sukhdev
  Singh}.} \bibinfo{year}{2022}\natexlab{}.
\newblock \showarticletitle{Natural language processing: State of the art,
  current trends and challenges}.
\newblock \bibinfo{journal}{\emph{Multimedia Tools and Applications}}
  (\bibinfo{year}{2022}), \bibinfo{pages}{1--32}.
\newblock


\bibitem[Kim and Mooney(2012)]%
        {DBLP:conf/emnlp/KimM12}
\bibfield{author}{\bibinfo{person}{Joohyun Kim} {and}
  \bibinfo{person}{Raymond~J. Mooney}.} \bibinfo{year}{2012}\natexlab{}.
\newblock \showarticletitle{Unsupervised {PCFG} Induction for Grounded Language
  Learning with Highly Ambiguous Supervision}. In
  \bibinfo{booktitle}{\emph{EMNLP-CoNLL}},
  \bibfield{editor}{\bibinfo{person}{Jun'ichi Tsujii}, \bibinfo{person}{James
  Henderson}, {and} \bibinfo{person}{Marius Pasca}} (Eds.).
  \bibinfo{pages}{433--444}.
\newblock
\urldef\tempurl%
\url{https://aclanthology.org/D12-1040/}
\showURL{%
\tempurl}


\bibitem[Koller and Hoffmann(2010)]%
        {DBLP:conf/aips/KollerH10}
\bibfield{author}{\bibinfo{person}{Alexander Koller} {and}
  \bibinfo{person}{J{\"{o}}rg Hoffmann}.} \bibinfo{year}{2010}\natexlab{}.
\newblock \showarticletitle{Waking Up a Sleeping Rabbit: On Natural-Language
  Sentence Generation with {FF}}. In \bibinfo{booktitle}{\emph{Proceedings of
  the 20th International Conference on Automated Planning and Scheduling,
  {ICAPS} 2010, Toronto, Ontario, Canada, May 12-16, 2010}},
  \bibfield{editor}{\bibinfo{person}{Ronen~I. Brafman}, \bibinfo{person}{Hector
  Geffner}, \bibinfo{person}{J{\"{o}}rg Hoffmann}, {and}
  \bibinfo{person}{Henry~A. Kautz}} (Eds.). \bibinfo{publisher}{{AAAI}},
  \bibinfo{pages}{238--241}.
\newblock
\urldef\tempurl%
\url{http://www.aaai.org/ocs/index.php/ICAPS/ICAPS10/paper/view/1415}
\showURL{%
\tempurl}


\bibitem[Kong et~al\mbox{.}(2021)]%
        {kong-etal-2021-stylized}
\bibfield{author}{\bibinfo{person}{Xiangzhe Kong}, \bibinfo{person}{Jialiang
  Huang}, \bibinfo{person}{Ziquan Tung}, \bibinfo{person}{Jian Guan}, {and}
  \bibinfo{person}{Minlie Huang}.} \bibinfo{year}{2021}\natexlab{}.
\newblock \showarticletitle{Stylized Story Generation with Style-Guided
  Planning}. In \bibinfo{booktitle}{\emph{ACL-IJCNLP}}.
  \bibinfo{pages}{2430--2436}.
\newblock


\bibitem[Kurisinkel et~al\mbox{.}(2017)]%
        {kurisinkel2017abstractive}
\bibfield{author}{\bibinfo{person}{Litton~J Kurisinkel}, \bibinfo{person}{Yue
  Zhang}, {and} \bibinfo{person}{Vasudeva Varma}.}
  \bibinfo{year}{2017}\natexlab{}.
\newblock \showarticletitle{Abstractive multi-document summarization by partial
  tree extraction, recombination and linearization}. In
  \bibinfo{booktitle}{\emph{Proceedings of the Eighth International Joint
  Conference on Natural Language Processing (Volume 1: Long Papers)}}.
  \bibinfo{pages}{812--821}.
\newblock


\bibitem[K{\"u}stenmacher and Pl{\"o}ger(2021)]%
        {kustenmacher2021improving}
\bibfield{author}{\bibinfo{person}{Anastassia K{\"u}stenmacher} {and}
  \bibinfo{person}{Paul~G Pl{\"o}ger}.} \bibinfo{year}{2021}\natexlab{}.
\newblock \showarticletitle{Improving the Reliability of Service Robots by
  Symbolic Representation of Execution Specific Knowledge}. In
  \bibinfo{booktitle}{\emph{Robust and Reliable Autonomy in the Wild (R2AW)}}.
\newblock


\bibitem[Li et~al\mbox{.}(2013)]%
        {DBLP:conf/aaai/LiLJR13}
\bibfield{author}{\bibinfo{person}{Boyang Li}, \bibinfo{person}{Stephen
  Lee{-}Urban}, \bibinfo{person}{George Johnston}, {and} \bibinfo{person}{Mark
  Riedl}.} \bibinfo{year}{2013}\natexlab{}.
\newblock \showarticletitle{Story Generation with Crowdsourced Plot Graphs}. In
  \bibinfo{booktitle}{\emph{Proceedings of the Twenty-Seventh {AAAI} Conference
  on Artificial Intelligence, July 14-18, 2013, Bellevue, Washington, {USA}}},
  \bibfield{editor}{\bibinfo{person}{Marie desJardins} {and}
  \bibinfo{person}{Michael~L. Littman}} (Eds.).
\newblock
\urldef\tempurl%
\url{http://www.aaai.org/ocs/index.php/AAAI/AAAI13/paper/view/6399}
\showURL{%
\tempurl}


\bibitem[Li et~al\mbox{.}(2022)]%
        {DBLP:journals/corr/abs-2202-01771}
\bibfield{author}{\bibinfo{person}{Shuang Li}, \bibinfo{person}{Xavier Puig},
  \bibinfo{person}{Chris Paxton}, \bibinfo{person}{Yilun Du},
  \bibinfo{person}{Clinton Wang}, \bibinfo{person}{Linxi Fan},
  \bibinfo{person}{Tao Chen}, \bibinfo{person}{De{-}An Huang},
  \bibinfo{person}{Ekin Aky{\"{u}}rek}, \bibinfo{person}{Anima Anandkumar},
  \bibinfo{person}{Jacob Andreas}, \bibinfo{person}{Igor Mordatch},
  \bibinfo{person}{Antonio Torralba}, {and} \bibinfo{person}{Yuke Zhu}.}
  \bibinfo{year}{2022}\natexlab{}.
\newblock \showarticletitle{Pre-Trained Language Models for Interactive
  Decision-Making}.
\newblock \bibinfo{journal}{\emph{CoRR}}  \bibinfo{volume}{abs/2202.01771}
  (\bibinfo{year}{2022}).
\newblock
\showeprint[arXiv]{2202.01771}
\urldef\tempurl%
\url{https://arxiv.org/abs/2202.01771}
\showURL{%
\tempurl}


\bibitem[Li et~al\mbox{.}(2021)]%
        {li2021terapipe}
\bibfield{author}{\bibinfo{person}{Zhuohan Li}, \bibinfo{person}{Siyuan
  Zhuang}, \bibinfo{person}{Shiyuan Guo}, \bibinfo{person}{Danyang Zhuo},
  \bibinfo{person}{Hao Zhang}, \bibinfo{person}{Dawn Song}, {and}
  \bibinfo{person}{Ion Stoica}.} \bibinfo{year}{2021}\natexlab{}.
\newblock \showarticletitle{Terapipe: Token-level pipeline parallelism for
  training large-scale language models}. In
  \bibinfo{booktitle}{\emph{International Conference on Machine Learning}}.
  PMLR, \bibinfo{pages}{6543--6552}.
\newblock


\bibitem[Lieven et~al\mbox{.}(2021)]%
        {lieven2021enabling}
\bibfield{author}{\bibinfo{person}{Claudius Lieven}, \bibinfo{person}{Bianca
  L{\"u}ders}, \bibinfo{person}{Daniel Kulus}, {and} \bibinfo{person}{Rosa
  Thoneick}.} \bibinfo{year}{2021}\natexlab{}.
\newblock \showarticletitle{Enabling digital co-creation in urban planning and
  development}.
\newblock  (\bibinfo{year}{2021}), \bibinfo{pages}{415--430}.
\newblock


\bibitem[Lindsay et~al\mbox{.}(2017)]%
        {DBLP:conf/aips/LindsayRFHPG17}
\bibfield{author}{\bibinfo{person}{Alan Lindsay}, \bibinfo{person}{Jonathon
  Read}, \bibinfo{person}{Jo{\~{a}}o~F. Ferreira}, \bibinfo{person}{Thomas
  Hayton}, \bibinfo{person}{Julie Porteous}, {and} \bibinfo{person}{Peter
  Gregory}.} \bibinfo{year}{2017}\natexlab{}.
\newblock \showarticletitle{Framer: Planning Models from Natural Language
  Action Descriptions}. In \bibinfo{booktitle}{\emph{{ICAPS}}}.
  \bibinfo{pages}{434--442}.
\newblock


\bibitem[Lopez(2008)]%
        {statisticMTsurvey}
\bibfield{author}{\bibinfo{person}{Adam Lopez}.}
  \bibinfo{year}{2008}\natexlab{}.
\newblock \showarticletitle{Statistical Machine Translation}.
\newblock \bibinfo{journal}{\emph{ACM Comput. Surv.}} \bibinfo{volume}{40},
  \bibinfo{number}{3}, Article \bibinfo{articleno}{8} (\bibinfo{date}{aug}
  \bibinfo{year}{2008}), \bibinfo{numpages}{49}~pages.
\newblock
\showISSN{0360-0300}
\urldef\tempurl%
\url{https://doi.org/10.1145/1380584.1380586}
\showDOI{\tempurl}


\bibitem[Lukin and Walker(2019)]%
        {lukin2019narrative}
\bibfield{author}{\bibinfo{person}{Stephanie~M Lukin} {and}
  \bibinfo{person}{Marilyn~A Walker}.} \bibinfo{year}{2019}\natexlab{}.
\newblock \showarticletitle{A narrative sentence planner and structurer for
  domain independent, parameterizable storytelling}.
\newblock \bibinfo{journal}{\emph{Dialogue \& Discourse}} \bibinfo{volume}{10},
  \bibinfo{number}{1} (\bibinfo{year}{2019}), \bibinfo{pages}{34--86}.
\newblock


\bibitem[Ma et~al\mbox{.}(2022)]%
        {DBLP:journals/tois/MaLZLL22}
\bibfield{author}{\bibinfo{person}{Longxuan Ma}, \bibinfo{person}{Mingda Li},
  \bibinfo{person}{Wei{-}Nan Zhang}, \bibinfo{person}{Jiapeng Li}, {and}
  \bibinfo{person}{Ting Liu}.} \bibinfo{year}{2022}\natexlab{}.
\newblock \showarticletitle{Unstructured Text Enhanced Open-Domain Dialogue
  System: {A} Systematic Survey}.
\newblock \bibinfo{journal}{\emph{{ACM} Trans. Inf. Syst.}}
  \bibinfo{volume}{40}, \bibinfo{number}{1} (\bibinfo{year}{2022}),
  \bibinfo{pages}{9:1--9:44}.
\newblock
\urldef\tempurl%
\url{https://doi.org/10.1145/3464377}
\showURL{%
\tempurl}


\bibitem[MacMahon et~al\mbox{.}(2006)]%
        {DBLP:conf/aaai/MacMahonSK06}
\bibfield{author}{\bibinfo{person}{Matt MacMahon}, \bibinfo{person}{Brian
  Stankiewicz}, {and} \bibinfo{person}{Benjamin Kuipers}.}
  \bibinfo{year}{2006}\natexlab{}.
\newblock \showarticletitle{Walk the Talk: Connecting Language, Knowledge, and
  Action in Route Instructions}. In \bibinfo{booktitle}{\emph{AAAI}}.
  \bibinfo{pages}{1475--1482}.
\newblock
\urldef\tempurl%
\url{http://www.aaai.org/Library/AAAI/2006/aaai06-232.php}
\showURL{%
\tempurl}


\bibitem[Marfurt and Henderson(2021)]%
        {marfurt2021sentence}
\bibfield{author}{\bibinfo{person}{Andreas Marfurt} {and}
  \bibinfo{person}{James Henderson}.} \bibinfo{year}{2021}\natexlab{}.
\newblock \showarticletitle{Sentence-level Planning for Especially Abstractive
  Summarization}. In \bibinfo{booktitle}{\emph{Proceedings of the Third
  Workshop on New Frontiers in Summarization}}. \bibinfo{pages}{1--14}.
\newblock


\bibitem[Matuszek et~al\mbox{.}(2010)]%
        {DBLP:conf/hri/MatuszekFK10}
\bibfield{author}{\bibinfo{person}{Cynthia Matuszek}, \bibinfo{person}{Dieter
  Fox}, {and} \bibinfo{person}{Karl Koscher}.} \bibinfo{year}{2010}\natexlab{}.
\newblock \showarticletitle{Following directions using statistical machine
  translation}. In \bibinfo{booktitle}{\emph{{HRI}}},
  \bibfield{editor}{\bibinfo{person}{Pamela~J. Hinds}, \bibinfo{person}{Hiroshi
  Ishiguro}, \bibinfo{person}{Takayuki Kanda}, {and} \bibinfo{person}{Peter
  H.~Kahn Jr.}} (Eds.). \bibinfo{pages}{251--258}.
\newblock
\urldef\tempurl%
\url{https://doi.org/10.1145/1734454.1734552}
\showURL{%
\tempurl}


\bibitem[McDermott et~al\mbox{.}(1998)]%
        {PDDL}
\bibfield{author}{\bibinfo{person}{Drew McDermott}, \bibinfo{person}{Malik
  Ghallab}, \bibinfo{person}{Adele~E. Howe}, \bibinfo{person}{Craig~A.
  Knoblock}, \bibinfo{person}{Ashwin Ram}, \bibinfo{person}{Manuela~M. Veloso},
  \bibinfo{person}{Daniel~S. Weld}, {and} \bibinfo{person}{David~E. Wilkins}.}
  \bibinfo{year}{1998}\natexlab{}.
\newblock \showarticletitle{PDDL-the planning domain definition language}.
\newblock


\bibitem[Mei et~al\mbox{.}(2016)]%
        {DBLP:conf/aaai/MeiBW16}
\bibfield{author}{\bibinfo{person}{Hongyuan Mei}, \bibinfo{person}{Mohit
  Bansal}, {and} \bibinfo{person}{Matthew~R. Walter}.}
  \bibinfo{year}{2016}\natexlab{}.
\newblock \showarticletitle{Listen, Attend, and Walk: Neural Mapping of
  Navigational Instructions to Action Sequences}. In
  \bibinfo{booktitle}{\emph{Thirtieth AAAI Conference on Artificial
  Intelligence}}, \bibfield{editor}{\bibinfo{person}{Dale Schuurmans} {and}
  \bibinfo{person}{Michael~P. Wellman}} (Eds.). \bibinfo{pages}{2772--2778}.
\newblock
\urldef\tempurl%
\url{http://www.aaai.org/ocs/index.php/AAAI/AAAI16/paper/view/12522}
\showURL{%
\tempurl}


\bibitem[Miah et~al\mbox{.}(2022)]%
        {miah2022social}
\bibfield{author}{\bibinfo{person}{Shah~Jahan Miah}, \bibinfo{person}{Huy~Quan
  Vu}, {and} \bibinfo{person}{Damminda Alahakoon}.}
  \bibinfo{year}{2022}\natexlab{}.
\newblock \showarticletitle{A social media analytics perspective for
  human-oriented smart city planning and management}.
\newblock \bibinfo{journal}{\emph{Journal of the Association for Information
  Science and Technology}} \bibinfo{volume}{73}, \bibinfo{number}{1}
  (\bibinfo{year}{2022}), \bibinfo{pages}{119--135}.
\newblock


\bibitem[Mohan et~al\mbox{.}(2016)]%
        {mohan2016study}
\bibfield{author}{\bibinfo{person}{M~Jishma Mohan}, \bibinfo{person}{C
  Sunitha}, \bibinfo{person}{Amal Ganesh}, {and} \bibinfo{person}{A Jaya}.}
  \bibinfo{year}{2016}\natexlab{}.
\newblock \showarticletitle{A study on ontology based abstractive
  summarization}.
\newblock \bibinfo{journal}{\emph{Procedia Computer Science}}
  \bibinfo{volume}{87} (\bibinfo{year}{2016}), \bibinfo{pages}{32--37}.
\newblock


\bibitem[Mohan and Laird(2014)]%
        {DBLP:conf/aaai/MohanL14}
\bibfield{author}{\bibinfo{person}{Shiwali Mohan} {and}
  \bibinfo{person}{John~E. Laird}.} \bibinfo{year}{2014}\natexlab{}.
\newblock \showarticletitle{Learning Goal-Oriented Hierarchical Tasks from
  Situated Interactive Instruction}. In \bibinfo{booktitle}{\emph{Proceedings
  of the Twenty-Eighth {AAAI} Conference on Artificial Intelligence, July 27
  -31, 2014, Qu{\'{e}}bec City, Qu{\'{e}}bec, Canada}},
  \bibfield{editor}{\bibinfo{person}{Carla~E. Brodley} {and}
  \bibinfo{person}{Peter Stone}} (Eds.). \bibinfo{pages}{387--394}.
\newblock
\urldef\tempurl%
\url{http://www.aaai.org/ocs/index.php/AAAI/AAAI14/paper/view/8630}
\showURL{%
\tempurl}


\bibitem[Muise et~al\mbox{.}(2019)]%
        {muise2019planning}
\bibfield{author}{\bibinfo{person}{Christian Muise}, \bibinfo{person}{Tathagata
  Chakraborti}, \bibinfo{person}{Shubham Agarwal}, \bibinfo{person}{Ondrej
  Bajgar}, \bibinfo{person}{Arunima Chaudhary}, \bibinfo{person}{Luis~A
  Lastras-Montano}, \bibinfo{person}{Josef Ondrej}, \bibinfo{person}{Miroslav
  Vodolan}, {and} \bibinfo{person}{Charlie Wiecha}.}
  \bibinfo{year}{2019}\natexlab{}.
\newblock \showarticletitle{Planning for goal-oriented dialogue systems}.
\newblock \bibinfo{journal}{\emph{arXiv preprint arXiv:1910.08137}}
  (\bibinfo{year}{2019}).
\newblock


\bibitem[Narayan et~al\mbox{.}(2021)]%
        {narayan2021planning}
\bibfield{author}{\bibinfo{person}{Shashi Narayan}, \bibinfo{person}{Yao Zhao},
  \bibinfo{person}{Joshua Maynez}, \bibinfo{person}{Gon{\c{c}}alo Sim{\~o}es},
  \bibinfo{person}{Vitaly Nikolaev}, {and} \bibinfo{person}{Ryan McDonald}.}
  \bibinfo{year}{2021}\natexlab{}.
\newblock \showarticletitle{Planning with learned entity prompts for
  abstractive summarization}.
\newblock \bibinfo{journal}{\emph{Transactions of the Association for
  Computational Linguistics}}  \bibinfo{volume}{9} (\bibinfo{year}{2021}),
  \bibinfo{pages}{1475--1492}.
\newblock


\bibitem[Narayanan et~al\mbox{.}(2021)]%
        {narayanan2021efficient}
\bibfield{author}{\bibinfo{person}{Deepak Narayanan}, \bibinfo{person}{Mohammad
  Shoeybi}, \bibinfo{person}{Jared Casper}, \bibinfo{person}{Patrick
  LeGresley}, \bibinfo{person}{Mostofa Patwary}, \bibinfo{person}{Vijay
  Korthikanti}, \bibinfo{person}{Dmitri Vainbrand}, \bibinfo{person}{Prethvi
  Kashinkunti}, \bibinfo{person}{Julie Bernauer}, \bibinfo{person}{Bryan
  Catanzaro}, {et~al\mbox{.}}} \bibinfo{year}{2021}\natexlab{}.
\newblock \showarticletitle{Efficient large-scale language model training on
  gpu clusters using megatron-lm}. In \bibinfo{booktitle}{\emph{Proceedings of
  the International Conference for High Performance Computing, Networking,
  Storage and Analysis}}. \bibinfo{pages}{1--15}.
\newblock


\bibitem[Nyga and Beetz(2012)]%
        {nyga2012everything}
\bibfield{author}{\bibinfo{person}{Daniel Nyga} {and} \bibinfo{person}{Michael
  Beetz}.} \bibinfo{year}{2012}\natexlab{}.
\newblock \showarticletitle{Everything robots always wanted to know about
  housework (but were afraid to ask)}. In \bibinfo{booktitle}{\emph{2012
  IEEE/RSJ International Conference on Intelligent Robots and Systems}}. IEEE,
  \bibinfo{pages}{243--250}.
\newblock


\bibitem[Okpor(2014)]%
        {okpor2014machine}
\bibfield{author}{\bibinfo{person}{MD Okpor}.} \bibinfo{year}{2014}\natexlab{}.
\newblock \showarticletitle{Machine translation approaches: issues and
  challenges}.
\newblock \bibinfo{journal}{\emph{International Journal of Computer Science
  Issues (IJCSI)}} \bibinfo{volume}{11}, \bibinfo{number}{5}
  (\bibinfo{year}{2014}), \bibinfo{pages}{159}.
\newblock


\bibitem[Otter et~al\mbox{.}(2020)]%
        {otter2020survey}
\bibfield{author}{\bibinfo{person}{Daniel~W Otter}, \bibinfo{person}{Julian~R
  Medina}, {and} \bibinfo{person}{Jugal~K Kalita}.}
  \bibinfo{year}{2020}\natexlab{}.
\newblock \showarticletitle{A survey of the usages of deep learning for natural
  language processing}.
\newblock \bibinfo{journal}{\emph{IEEE transactions on neural networks and
  learning systems}} \bibinfo{volume}{32}, \bibinfo{number}{2}
  (\bibinfo{year}{2020}), \bibinfo{pages}{604--624}.
\newblock


\bibitem[Pallagani and Srivastava(2021)]%
        {pallagani2021generic}
\bibfield{author}{\bibinfo{person}{Vishal Pallagani} {and}
  \bibinfo{person}{Biplav Srivastava}.} \bibinfo{year}{2021}\natexlab{}.
\newblock \showarticletitle{A Generic Dialog Agent for Information Retrieval
  Based on Automated Planning Within a Reinforcement Learning Platform}.
\newblock \bibinfo{journal}{\emph{Bridging the Gap Between AI Planning and
  Reinforcement Learning (PRL)}} (\bibinfo{year}{2021}).
\newblock


\bibitem[Perrault and Allen(1980)]%
        {DBLP:journals/coling/PerraultA80}
\bibfield{author}{\bibinfo{person}{C.~Raymond Perrault} {and}
  \bibinfo{person}{James~F. Allen}.} \bibinfo{year}{1980}\natexlab{}.
\newblock \showarticletitle{A Plan-Based Analysis of Indirect Speech Acts}.
\newblock \bibinfo{journal}{\emph{Am. J. Comput. Linguistics}}
  \bibinfo{volume}{6}, \bibinfo{number}{3-4} (\bibinfo{year}{1980}),
  \bibinfo{pages}{167--182}.
\newblock


\bibitem[Petrick and Foster(2016)]%
        {petrick2016using}
\bibfield{author}{\bibinfo{person}{Ronald~PA Petrick} {and}
  \bibinfo{person}{Mary~Ellen Foster}.} \bibinfo{year}{2016}\natexlab{}.
\newblock \showarticletitle{Using general-purpose planning for action selection
  in human-robot interaction}. In \bibinfo{booktitle}{\emph{2016 AAAI Fall
  Symposium Series}}.
\newblock


\bibitem[Pham and Yoshimi(2015)]%
        {pham2015extraction}
\bibfield{author}{\bibinfo{person}{Ngoc~Hung Pham} {and}
  \bibinfo{person}{Takashi Yoshimi}.} \bibinfo{year}{2015}\natexlab{}.
\newblock \showarticletitle{Extraction of actions and objects from instruction
  manual for executable robot planning}. In \bibinfo{booktitle}{\emph{2015 15th
  International Conference on Control, Automation and Systems (ICCAS)}}. IEEE,
  \bibinfo{pages}{881--885}.
\newblock


\bibitem[Porteous and Cavazza(2009)]%
        {DBLP:conf/icids/PorteousC09}
\bibfield{author}{\bibinfo{person}{Julie Porteous} {and} \bibinfo{person}{Marc
  Cavazza}.} \bibinfo{year}{2009}\natexlab{}.
\newblock \showarticletitle{Controlling Narrative Generation with Planning
  Trajectories: The Role of Constraints}. In \bibinfo{booktitle}{\emph{{ICIDS}
  2009}} \emph{(\bibinfo{series}{Lecture Notes in Computer Science},
  Vol.~\bibinfo{volume}{5915})}, \bibfield{editor}{\bibinfo{person}{Ido
  Iurgel}, \bibinfo{person}{Nelson Zagalo}, {and} \bibinfo{person}{Paolo
  Petta}} (Eds.). \bibinfo{pages}{234--245}.
\newblock
\urldef\tempurl%
\url{https://doi.org/10.1007/978-3-642-10643-9\_28}
\showURL{%
\tempurl}


\bibitem[Porteous et~al\mbox{.}(2021a)]%
        {DBLP:journals/aamas/PorteousFLC21}
\bibfield{author}{\bibinfo{person}{Julie Porteous},
  \bibinfo{person}{Jo{\~{a}}o~F. Ferreira}, \bibinfo{person}{Alan Lindsay},
  {and} \bibinfo{person}{Marc Cavazza}.} \bibinfo{year}{2021}\natexlab{a}.
\newblock \showarticletitle{Automated narrative planning model extension}.
\newblock \bibinfo{journal}{\emph{Auton. Agents Multi Agent Syst.}}
  \bibinfo{volume}{35}, \bibinfo{number}{2} (\bibinfo{year}{2021}),
  \bibinfo{pages}{19}.
\newblock
\urldef\tempurl%
\url{https://doi.org/10.1007/s10458-021-09501-1}
\showURL{%
\tempurl}


\bibitem[Porteous et~al\mbox{.}(2021b)]%
        {porteous2021automated}
\bibfield{author}{\bibinfo{person}{Julie Porteous}, \bibinfo{person}{Jo{\~a}o~F
  Ferreira}, \bibinfo{person}{Alan Lindsay}, {and} \bibinfo{person}{Marc
  Cavazza}.} \bibinfo{year}{2021}\natexlab{b}.
\newblock \showarticletitle{Automated narrative planning model extension}.
\newblock \bibinfo{journal}{\emph{Autonomous Agents and Multi-Agent Systems}}
  \bibinfo{volume}{35}, \bibinfo{number}{2} (\bibinfo{year}{2021}),
  \bibinfo{pages}{1--29}.
\newblock


\bibitem[Resch et~al\mbox{.}(2016)]%
        {resch2016citizen}
\bibfield{author}{\bibinfo{person}{Bernd Resch}, \bibinfo{person}{Anja Summa},
  \bibinfo{person}{Peter Zeile}, {and} \bibinfo{person}{Michael Strube}.}
  \bibinfo{year}{2016}\natexlab{}.
\newblock \showarticletitle{Citizen-centric urban planning through extracting
  emotion information from twitter in an interdisciplinary
  space-time-linguistics algorithm}.
\newblock \bibinfo{journal}{\emph{Urban Planning}} \bibinfo{volume}{1},
  \bibinfo{number}{2} (\bibinfo{year}{2016}), \bibinfo{pages}{114--127}.
\newblock


\bibitem[Riedl and Young(2010)]%
        {DBLP:journals/jair/RiedlY10_IPOCL}
\bibfield{author}{\bibinfo{person}{Mark~O. Riedl} {and}
  \bibinfo{person}{Robert~Michael Young}.} \bibinfo{year}{2010}\natexlab{}.
\newblock \showarticletitle{Narrative Planning: Balancing Plot and Character}.
\newblock \bibinfo{journal}{\emph{J. Artif. Intell. Res.}}
  \bibinfo{volume}{39} (\bibinfo{year}{2010}), \bibinfo{pages}{217--268}.
\newblock
\urldef\tempurl%
\url{https://doi.org/10.1613/jair.2989}
\showURL{%
\tempurl}


\bibitem[Rose et~al\mbox{.}(2010)]%
        {rose2010automatic}
\bibfield{author}{\bibinfo{person}{Stuart Rose}, \bibinfo{person}{Dave Engel},
  \bibinfo{person}{Nick Cramer}, {and} \bibinfo{person}{Wendy Cowley}.}
  \bibinfo{year}{2010}\natexlab{}.
\newblock \showarticletitle{Automatic keyword extraction from individual
  documents}.
\newblock \bibinfo{journal}{\emph{Text mining: applications and theory}}
  \bibinfo{volume}{1} (\bibinfo{year}{2010}), \bibinfo{pages}{1--20}.
\newblock


\bibitem[Sanner et~al\mbox{.}(2010)]%
        {rddl}
\bibfield{author}{\bibinfo{person}{Scott Sanner} {et~al\mbox{.}}}
  \bibinfo{year}{2010}\natexlab{}.
\newblock \showarticletitle{Relational dynamic influence diagram language
  (rddl): Language description}.
\newblock \bibinfo{journal}{\emph{Unpublished ms. Australian National
  University}}  \bibinfo{volume}{32} (\bibinfo{year}{2010}),
  \bibinfo{pages}{27}.
\newblock


\bibitem[Santos~Teixeira and Dragoni(2022)]%
        {santos2022review}
\bibfield{author}{\bibinfo{person}{Milene Santos~Teixeira} {and}
  \bibinfo{person}{Mauro Dragoni}.} \bibinfo{year}{2022}\natexlab{}.
\newblock \showarticletitle{A Review of Plan-Based Approaches for Dialogue
  Management}.
\newblock \bibinfo{journal}{\emph{Cognitive Computation}}
  (\bibinfo{year}{2022}), \bibinfo{pages}{1--20}.
\newblock


\bibitem[Say(2021)]%
        {DBLP:conf/aaai/Say21}
\bibfield{author}{\bibinfo{person}{Buser Say}.}
  \bibinfo{year}{2021}\natexlab{}.
\newblock \showarticletitle{A Unified Framework for Planning with Learned
  Neural Network Transition Models}.
\newblock  (\bibinfo{year}{2021}), \bibinfo{pages}{5016--5024}.
\newblock
\urldef\tempurl%
\url{https://ojs.aaai.org/index.php/AAAI/article/view/16635}
\showURL{%
\tempurl}


\bibitem[Scheutz et~al\mbox{.}(2017)]%
        {DBLP:conf/atal/ScheutzKOFP17}
\bibfield{author}{\bibinfo{person}{Matthias Scheutz}, \bibinfo{person}{Evan~A.
  Krause}, \bibinfo{person}{Bradley Oosterveld}, \bibinfo{person}{Tyler~M.
  Frasca}, {and} \bibinfo{person}{Robert~Platt Jr.}}
  \bibinfo{year}{2017}\natexlab{}.
\newblock \showarticletitle{Spoken Instruction-Based One-Shot Object and Action
  Learning in a Cognitive Robotic Architecture}. In
  \bibinfo{booktitle}{\emph{Proceedings of the 16th Conference on Autonomous
  Agents and MultiAgent Systems, {AAMAS} 2017, S{\~{a}}o Paulo, Brazil, May
  8-12, 2017}}, \bibfield{editor}{\bibinfo{person}{Kate Larson},
  \bibinfo{person}{Michael Winikoff}, \bibinfo{person}{Sanmay Das}, {and}
  \bibinfo{person}{Edmund~H. Durfee}} (Eds.). \bibinfo{pages}{1378--1386}.
\newblock
\urldef\tempurl%
\url{http://dl.acm.org/citation.cfm?id=3091315}
\showURL{%
\tempurl}


\bibitem[Sha et~al\mbox{.}(2018)]%
        {DBLP:conf/aaai/ShaMLPLCS18}
\bibfield{author}{\bibinfo{person}{Lei Sha}, \bibinfo{person}{Lili Mou},
  \bibinfo{person}{Tianyu Liu}, \bibinfo{person}{Pascal Poupart},
  \bibinfo{person}{Sujian Li}, \bibinfo{person}{Baobao Chang}, {and}
  \bibinfo{person}{Zhifang Sui}.} \bibinfo{year}{2018}\natexlab{}.
\newblock \showarticletitle{Order-Planning Neural Text Generation From
  Structured Data}. In \bibinfo{booktitle}{\emph{AAAI}},
  \bibfield{editor}{\bibinfo{person}{Sheila~A. McIlraith} {and}
  \bibinfo{person}{Kilian~Q. Weinberger}} (Eds.). \bibinfo{pages}{5414--5421}.
\newblock


\bibitem[Shaalan et~al\mbox{.}(2010)]%
        {shaalan2010rule}
\bibfield{author}{\bibinfo{person}{Khaled Shaalan} {et~al\mbox{.}}}
  \bibinfo{year}{2010}\natexlab{}.
\newblock \showarticletitle{Rule-based approach in Arabic natural language
  processing}.
\newblock \bibinfo{journal}{\emph{The International Journal on Information and
  Communication Technologies (IJICT)}} \bibinfo{volume}{3}, \bibinfo{number}{3}
  (\bibinfo{year}{2010}), \bibinfo{pages}{11--19}.
\newblock


\bibitem[Sharma et~al\mbox{.}(2022)]%
        {DBLP:conf/acl/Sharma0A22}
\bibfield{author}{\bibinfo{person}{Pratyusha Sharma}, \bibinfo{person}{Antonio
  Torralba}, {and} \bibinfo{person}{Jacob Andreas}.}
  \bibinfo{year}{2022}\natexlab{}.
\newblock \showarticletitle{Skill Induction and Planning with Latent Language}.
\newblock  (\bibinfo{year}{2022}), \bibinfo{pages}{1713--1726}.
\newblock
\urldef\tempurl%
\url{https://doi.org/10.18653/v1/2022.acl-long.120}
\showURL{%
\tempurl}


\bibitem[She and Chai(2017)]%
        {DBLP:conf/acl/SheC17}
\bibfield{author}{\bibinfo{person}{Lanbo She} {and} \bibinfo{person}{Joyce~Yue
  Chai}.} \bibinfo{year}{2017}\natexlab{}.
\newblock \showarticletitle{Interactive Learning of Grounded Verb Semantics
  towards Human-Robot Communication}. In \bibinfo{booktitle}{\emph{Proceedings
  of the 55th Annual Meeting of the Association for Computational Linguistics,
  {ACL} 2017, Vancouver, Canada, July 30 - August 4, Volume 1: Long Papers}},
  \bibfield{editor}{\bibinfo{person}{Regina Barzilay} {and}
  \bibinfo{person}{Min{-}Yen Kan}} (Eds.). \bibinfo{pages}{1634--1644}.
\newblock
\urldef\tempurl%
\url{https://doi.org/10.18653/v1/P17-1150}
\showURL{%
\tempurl}


\bibitem[She et~al\mbox{.}(2014a)]%
        {DBLP:conf/ro-man/SheCCJYX14}
\bibfield{author}{\bibinfo{person}{Lanbo She}, \bibinfo{person}{Yu Cheng},
  \bibinfo{person}{Joyce~Yue Chai}, \bibinfo{person}{Yunyi Jia},
  \bibinfo{person}{Shaohua Yang}, {and} \bibinfo{person}{Ning Xi}.}
  \bibinfo{year}{2014}\natexlab{a}.
\newblock \showarticletitle{Teaching Robots New Actions through Natural
  Language Instructions}. In \bibinfo{booktitle}{\emph{The 23rd {IEEE}
  International Symposium on Robot and Human Interactive Communication, {IEEE}
  {RO-MAN} 2014, Edinburgh, UK, August 25-29, 2014}}.
  \bibinfo{pages}{868--873}.
\newblock
\urldef\tempurl%
\url{https://doi.org/10.1109/ROMAN.2014.6926362}
\showURL{%
\tempurl}


\bibitem[She et~al\mbox{.}(2014b)]%
        {DBLP:conf/sigdial/SheYCJCX14}
\bibfield{author}{\bibinfo{person}{Lanbo She}, \bibinfo{person}{Shaohua Yang},
  \bibinfo{person}{Yu Cheng}, \bibinfo{person}{Yunyi Jia},
  \bibinfo{person}{Joyce~Yue Chai}, {and} \bibinfo{person}{Ning Xi}.}
  \bibinfo{year}{2014}\natexlab{b}.
\newblock \showarticletitle{Back to the Blocks World: Learning New Actions
  through Situated Human-Robot Dialogue}. In
  \bibinfo{booktitle}{\emph{Proceedings of the {SIGDIAL} 2014 Conference, The
  15th Annual Meeting of the Special Interest Group on Discourse and Dialogue,
  18-20 June 2014, Philadelphia, PA, {USA}}}. \bibinfo{pages}{89--97}.
\newblock
\urldef\tempurl%
\url{https://doi.org/10.3115/v1/w14-4313}
\showURL{%
\tempurl}


\bibitem[Shu and Nakayama(2018)]%
        {DBLP:journals/corr/abs-1808-04525}
\bibfield{author}{\bibinfo{person}{Raphael Shu} {and} \bibinfo{person}{Hideki
  Nakayama}.} \bibinfo{year}{2018}\natexlab{}.
\newblock \showarticletitle{Discrete Structural Planning for Neural Machine
  Translation}.
\newblock \bibinfo{journal}{\emph{CoRR}}  \bibinfo{volume}{abs/1808.04525}
  (\bibinfo{year}{2018}).
\newblock
\showeprint[arXiv]{1808.04525}
\urldef\tempurl%
\url{http://arxiv.org/abs/1808.04525}
\showURL{%
\tempurl}


\bibitem[Sil and Yates(2011)]%
        {DBLP:conf/ranlp/SilY11}
\bibfield{author}{\bibinfo{person}{Avirup Sil} {and} \bibinfo{person}{Alexander
  Yates}.} \bibinfo{year}{2011}\natexlab{}.
\newblock \showarticletitle{Extracting {STRIPS} Representations of Actions and
  Events}. In \bibinfo{booktitle}{\emph{{RANLP}}}. \bibinfo{pages}{1--8}.
\newblock


\bibitem[Simon and Muise(2022)]%
        {simon2022tattletale}
\bibfield{author}{\bibinfo{person}{Nisha Simon} {and}
  \bibinfo{person}{Christian Muise}.} \bibinfo{year}{2022}\natexlab{}.
\newblock \showarticletitle{TattleTale: Storytelling with Planning and Large
  Language Models}.
\newblock  (\bibinfo{year}{2022}).
\newblock


\bibitem[Sreedharan et~al\mbox{.}(2020)]%
        {sreedharan2020d3wa+}
\bibfield{author}{\bibinfo{person}{Sarath Sreedharan},
  \bibinfo{person}{Tathagata Chakraborti}, \bibinfo{person}{Christian Muise},
  \bibinfo{person}{Yasaman Khazaeni}, {and} \bibinfo{person}{Subbarao
  Kambhampati}.} \bibinfo{year}{2020}\natexlab{}.
\newblock \showarticletitle{--d3wa+--a case study of xaip in a model
  acquisition task for dialogue planning}. In
  \bibinfo{booktitle}{\emph{Proceedings of the International Conference on
  Automated Planning and Scheduling}}, Vol.~\bibinfo{volume}{30}.
  \bibinfo{pages}{488--497}.
\newblock


\bibitem[Stahlberg(2020)]%
        {DBLP:journals/jair/Stahlberg20}
\bibfield{author}{\bibinfo{person}{Felix Stahlberg}.}
  \bibinfo{year}{2020}\natexlab{}.
\newblock \showarticletitle{Neural Machine Translation: {A} Review}.
\newblock \bibinfo{journal}{\emph{J. Artif. Intell. Res.}}
  \bibinfo{volume}{69} (\bibinfo{year}{2020}), \bibinfo{pages}{343--418}.
\newblock
\urldef\tempurl%
\url{https://doi.org/10.1613/jair.1.12007}
\showURL{%
\tempurl}


\bibitem[Storks et~al\mbox{.}(2019)]%
        {storks2019commonsense}
\bibfield{author}{\bibinfo{person}{Shane Storks}, \bibinfo{person}{Qiaozi Gao},
  {and} \bibinfo{person}{Joyce~Y Chai}.} \bibinfo{year}{2019}\natexlab{}.
\newblock \showarticletitle{Commonsense reasoning for natural language
  understanding: A survey of benchmarks, resources, and approaches}.
\newblock \bibinfo{journal}{\emph{arXiv preprint arXiv:1904.01172}}
  (\bibinfo{year}{2019}), \bibinfo{pages}{1--60}.
\newblock


\bibitem[Suddrey et~al\mbox{.}(2022)]%
        {suddrey2022learning}
\bibfield{author}{\bibinfo{person}{Gavin Suddrey}, \bibinfo{person}{Ben
  Talbot}, {and} \bibinfo{person}{Frederic Maire}.}
  \bibinfo{year}{2022}\natexlab{}.
\newblock \showarticletitle{Learning and executing re-usable behaviour trees
  from natural language instruction}.
\newblock \bibinfo{journal}{\emph{IEEE Robotics and Automation Letters}}
  (\bibinfo{year}{2022}).
\newblock


\bibitem[Tambwekar et~al\mbox{.}(2019)]%
        {DBLP:conf/ijcai/TambwekarDMMHR19}
\bibfield{author}{\bibinfo{person}{Pradyumna Tambwekar},
  \bibinfo{person}{Murtaza Dhuliawala}, \bibinfo{person}{Lara~J. Martin},
  \bibinfo{person}{Animesh Mehta}, \bibinfo{person}{Brent Harrison}, {and}
  \bibinfo{person}{Mark~O. Riedl}.} \bibinfo{year}{2019}\natexlab{}.
\newblock \showarticletitle{Controllable Neural Story Plot Generation via
  Reward Shaping}. In \bibinfo{booktitle}{\emph{Proceedings of the
  Twenty-Eighth International Joint Conference on Artificial Intelligence,
  {IJCAI} 2019, Macao, China, August 10-16, 2019}},
  \bibfield{editor}{\bibinfo{person}{Sarit Kraus}} (Ed.).
  \bibinfo{pages}{5982--5988}.
\newblock
\urldef\tempurl%
\url{https://doi.org/10.24963/ijcai.2019/829}
\showURL{%
\tempurl}


\bibitem[Tellex et~al\mbox{.}(2011)]%
        {DBLP:conf/aaai/TellexKDWBTR11}
\bibfield{author}{\bibinfo{person}{Stefanie Tellex}, \bibinfo{person}{Thomas
  Kollar}, \bibinfo{person}{Steven Dickerson}, \bibinfo{person}{Matthew~R.
  Walter}, \bibinfo{person}{Ashis~Gopal Banerjee}, \bibinfo{person}{Seth~J.
  Teller}, {and} \bibinfo{person}{Nicholas Roy}.}
  \bibinfo{year}{2011}\natexlab{}.
\newblock \showarticletitle{Understanding Natural Language Commands for Robotic
  Navigation and Mobile Manipulation}. In \bibinfo{booktitle}{\emph{{AAAI}}},
  \bibfield{editor}{\bibinfo{person}{Wolfram Burgard} {and}
  \bibinfo{person}{Dan Roth}} (Eds.).
\newblock
\urldef\tempurl%
\url{http://www.aaai.org/ocs/index.php/AAAI/AAAI11/paper/view/3623}
\showURL{%
\tempurl}


\bibitem[Tenorth et~al\mbox{.}(2010)]%
        {DBLP:conf/icra/TenorthNB10}
\bibfield{author}{\bibinfo{person}{Moritz Tenorth}, \bibinfo{person}{Daniel
  Nyga}, {and} \bibinfo{person}{Michael Beetz}.}
  \bibinfo{year}{2010}\natexlab{}.
\newblock \showarticletitle{Understanding and executing instructions for
  everyday manipulation tasks from the World Wide Web}. In
  \bibinfo{booktitle}{\emph{{IEEE} International Conference on Robotics and
  Automation, {ICRA} 2010, Anchorage, Alaska, USA, 3-7 May 2010}}.
  \bibinfo{pages}{1486--1491}.
\newblock
\urldef\tempurl%
\url{https://doi.org/10.1109/ROBOT.2010.5509955}
\showURL{%
\tempurl}


\bibitem[Thomason et~al\mbox{.}(2019)]%
        {DBLP:conf/corl/ThomasonMCZ19}
\bibfield{author}{\bibinfo{person}{Jesse Thomason}, \bibinfo{person}{Michael
  Murray}, \bibinfo{person}{Maya Cakmak}, {and} \bibinfo{person}{Luke
  Zettlemoyer}.} \bibinfo{year}{2019}\natexlab{}.
\newblock \showarticletitle{Vision-and-Dialog Navigation}. In
  \bibinfo{booktitle}{\emph{3rd Annual Conference on Robot Learning, CoRL 2019,
  Osaka, Japan, October 30 - November 1, 2019, Proceedings}}
  \emph{(\bibinfo{series}{Proceedings of Machine Learning Research},
  Vol.~\bibinfo{volume}{100})}, \bibfield{editor}{\bibinfo{person}{Leslie~Pack
  Kaelbling}, \bibinfo{person}{Danica Kragic}, {and} \bibinfo{person}{Komei
  Sugiura}} (Eds.). \bibinfo{pages}{394--406}.
\newblock
\urldef\tempurl%
\url{http://proceedings.mlr.press/v100/thomason20a.html}
\showURL{%
\tempurl}


\bibitem[Thomason et~al\mbox{.}(2015)]%
        {DBLP:conf/ijcai/ThomasonZMS15}
\bibfield{author}{\bibinfo{person}{Jesse Thomason}, \bibinfo{person}{Shiqi
  Zhang}, \bibinfo{person}{Raymond~J. Mooney}, {and} \bibinfo{person}{Peter
  Stone}.} \bibinfo{year}{2015}\natexlab{}.
\newblock \showarticletitle{Learning to Interpret Natural Language Commands
  through Human-Robot Dialog}. In \bibinfo{booktitle}{\emph{{IJCAI} 2015}},
  \bibfield{editor}{\bibinfo{person}{Qiang Yang} {and}
  \bibinfo{person}{Michael~J. Wooldridge}} (Eds.). \bibinfo{pages}{1923--1929}.
\newblock
\urldef\tempurl%
\url{http://ijcai.org/Abstract/15/273}
\showURL{%
\tempurl}


\bibitem[Torfi et~al\mbox{.}(2020)]%
        {torfi2020natural}
\bibfield{author}{\bibinfo{person}{Amirsina Torfi}, \bibinfo{person}{Rouzbeh~A
  Shirvani}, \bibinfo{person}{Yaser Keneshloo}, \bibinfo{person}{Nader Tavaf},
  {and} \bibinfo{person}{Edward~A Fox}.} \bibinfo{year}{2020}\natexlab{}.
\newblock \showarticletitle{Natural language processing advancements by deep
  learning: A survey}.
\newblock \bibinfo{journal}{\emph{arXiv preprint arXiv:2003.01200}}
  (\bibinfo{year}{2020}).
\newblock


\bibitem[Ware and Young(2011)]%
        {DBLP:conf/aiide/WareY11_CPOCL}
\bibfield{author}{\bibinfo{person}{Stephen~G. Ware} {and}
  \bibinfo{person}{Robert~Michael Young}.} \bibinfo{year}{2011}\natexlab{}.
\newblock \showarticletitle{{CPOCL:} {A} Narrative Planner Supporting
  Conflict}. In \bibinfo{booktitle}{\emph{{AIIDE}}},
  \bibfield{editor}{\bibinfo{person}{Vadim Bulitko} {and}
  \bibinfo{person}{Mark~O. Riedl}} (Eds.).
\newblock
\urldef\tempurl%
\url{http://www.aaai.org/ocs/index.php/AIIDE/AIIDE11/paper/view/4058}
\showURL{%
\tempurl}


\bibitem[Wilensky(1981)]%
        {DBLP:journals/cogsci/Wilensky81}
\bibfield{author}{\bibinfo{person}{Robert Wilensky}.}
  \bibinfo{year}{1981}\natexlab{}.
\newblock \showarticletitle{Meta-Planning: Representing and Using Knowledge
  About Planning in Problem Solving and Natural Language Understanding}.
\newblock \bibinfo{journal}{\emph{Cogn. Sci.}} \bibinfo{volume}{5},
  \bibinfo{number}{3} (\bibinfo{year}{1981}), \bibinfo{pages}{197--233}.
\newblock
\urldef\tempurl%
\url{https://doi.org/10.1207/s15516709cog0503\_2}
\showURL{%
\tempurl}


\bibitem[Xiang et~al\mbox{.}(2015)]%
        {DBLP:conf/emnlp/XiangJCS15}
\bibfield{author}{\bibinfo{person}{Chuncheng Xiang}, \bibinfo{person}{Tingsong
  Jiang}, \bibinfo{person}{Baobao Chang}, {and} \bibinfo{person}{Zhifang Sui}.}
  \bibinfo{year}{2015}\natexlab{}.
\newblock \showarticletitle{{ERSOM:} {A} Structural Ontology Matching Approach
  Using Automatically Learned Entity Representation}. In
  \bibinfo{booktitle}{\emph{Proceedings of the 2015 Conference on Empirical
  Methods in Natural Language Processing, {EMNLP} 2015, Lisbon, Portugal,
  September 17-21, 2015}}, \bibfield{editor}{\bibinfo{person}{Llu{\'{\i}}s
  M{\`{a}}rquez}, \bibinfo{person}{Chris Callison{-}Burch},
  \bibinfo{person}{Jian Su}, \bibinfo{person}{Daniele Pighin}, {and}
  \bibinfo{person}{Yuval Marton}} (Eds.). \bibinfo{pages}{2419--2429}.
\newblock
\urldef\tempurl%
\url{https://doi.org/10.18653/v1/d15-1289}
\showURL{%
\tempurl}


\bibitem[Xu et~al\mbox{.}(2018)]%
        {DBLP:conf/emnlp/XuRZZC018}
\bibfield{author}{\bibinfo{person}{Jingjing Xu}, \bibinfo{person}{Xuancheng
  Ren}, \bibinfo{person}{Yi Zhang}, \bibinfo{person}{Qi Zeng},
  \bibinfo{person}{Xiaoyan Cai}, {and} \bibinfo{person}{Xu Sun}.}
  \bibinfo{year}{2018}\natexlab{}.
\newblock \showarticletitle{A Skeleton-Based Model for Promoting Coherence
  Among Sentences in Narrative Story Generation}. In
  \bibinfo{booktitle}{\emph{{EMNLP}}}. \bibinfo{pages}{4306--4315}.
\newblock
\urldef\tempurl%
\url{https://aclanthology.org/D18-1462/}
\showURL{%
\tempurl}


\bibitem[Yao et~al\mbox{.}(2019)]%
        {DBLP:conf/aaai/YaoPWK0Y19}
\bibfield{author}{\bibinfo{person}{Lili Yao}, \bibinfo{person}{Nanyun Peng},
  \bibinfo{person}{Ralph~M. Weischedel}, \bibinfo{person}{Kevin Knight},
  \bibinfo{person}{Dongyan Zhao}, {and} \bibinfo{person}{Rui Yan}.}
  \bibinfo{year}{2019}\natexlab{}.
\newblock \showarticletitle{Plan-and-Write: Towards Better Automatic
  Storytelling}. In \bibinfo{booktitle}{\emph{{AAAI}}}.
  \bibinfo{pages}{7378--7385}.
\newblock


\bibitem[Ye et~al\mbox{.}(2021)]%
        {DBLP:conf/robio/YeXLHSCS21}
\bibfield{author}{\bibinfo{person}{Rongguang Ye}, \bibinfo{person}{Qingchuan
  Xu}, \bibinfo{person}{Jie Liu}, \bibinfo{person}{Yang Hong},
  \bibinfo{person}{Chengfeng Sun}, \bibinfo{person}{Wenzheng Chi}, {and}
  \bibinfo{person}{Lining Sun}.} \bibinfo{year}{2021}\natexlab{}.
\newblock \showarticletitle{A Natural Language Instruction Disambiguation
  Method for Robot Grasping}. In \bibinfo{booktitle}{\emph{{IEEE} International
  Conference on Robotics and Biomimetics, {ROBIO} 2021, Sanya, China, December
  27-31, 2021}}. \bibinfo{pages}{601--606}.
\newblock
\urldef\tempurl%
\url{https://doi.org/10.1109/ROBIO54168.2021.9739456}
\showURL{%
\tempurl}


\bibitem[Yordanova and Kirste(2016)]%
        {DBLP:conf/icaart/YordanovaK16}
\bibfield{author}{\bibinfo{person}{Kristina~Y. Yordanova} {and}
  \bibinfo{person}{Thomas Kirste}.} \bibinfo{year}{2016}\natexlab{}.
\newblock \showarticletitle{Learning Models of Human Behaviour from Textual
  Instructions}. In \bibinfo{booktitle}{\emph{{ICAART}}}.
  \bibinfo{pages}{415--422}.
\newblock


\bibitem[Young et~al\mbox{.}(2013)]%
        {young2013plans}
\bibfield{author}{\bibinfo{person}{R~Michael Young}, \bibinfo{person}{Stephen~G
  Ware}, \bibinfo{person}{Brad~A Cassell}, {and} \bibinfo{person}{Justus
  Robertson}.} \bibinfo{year}{2013}\natexlab{}.
\newblock \showarticletitle{Plans and planning in narrative generation: a
  review of plan-based approaches to the generation of story, discourse and
  interactivity in narratives}.
\newblock \bibinfo{journal}{\emph{Sprache und Datenverarbeitung, Special Issue
  on Formal and Computational Models of Narrative}} \bibinfo{volume}{37},
  \bibinfo{number}{1-2} (\bibinfo{year}{2013}), \bibinfo{pages}{41--64}.
\newblock


\bibitem[Yu et~al\mbox{.}(2021)]%
        {DBLP:conf/aaai/YuLCY021}
\bibfield{author}{\bibinfo{person}{Meng{-}Hsuan Yu}, \bibinfo{person}{Juntao
  Li}, \bibinfo{person}{Zhangming Chan}, \bibinfo{person}{Rui Yan}, {and}
  \bibinfo{person}{Dongyan Zhao}.} \bibinfo{year}{2021}\natexlab{}.
\newblock \showarticletitle{Content Learning with Structure-Aware Writing: {A}
  Graph-Infused Dual Conditional Variational Autoencoder for Automatic
  Storytelling}. In \bibinfo{booktitle}{\emph{{AAAI}}}.
  \bibinfo{pages}{6021--6029}.
\newblock


\bibitem[Yu et~al\mbox{.}(2022)]%
        {yu2022survey}
\bibfield{author}{\bibinfo{person}{Wenhao Yu}, \bibinfo{person}{Chenguang Zhu},
  \bibinfo{person}{Zaitang Li}, \bibinfo{person}{Zhiting Hu},
  \bibinfo{person}{Qingyun Wang}, \bibinfo{person}{Heng Ji}, {and}
  \bibinfo{person}{Meng Jiang}.} \bibinfo{year}{2022}\natexlab{}.
\newblock \showarticletitle{A survey of knowledge-enhanced text generation}.
\newblock \bibinfo{journal}{\emph{ACM Computing Surveys (CSUR)}}
  (\bibinfo{year}{2022}).
\newblock


\bibitem[Zha et~al\mbox{.}(2021)]%
        {DBLP:journals/corr/abs-2110-05286}
\bibfield{author}{\bibinfo{person}{Yantian Zha}, \bibinfo{person}{Lin Guan},
  {and} \bibinfo{person}{Subbarao Kambhampati}.}
  \bibinfo{year}{2021}\natexlab{}.
\newblock \showarticletitle{Learning from Ambiguous Demonstrations with
  Self-Explanation Guided Reinforcement Learning}.
\newblock \bibinfo{journal}{\emph{CoRR}}  \bibinfo{volume}{abs/2110.05286}
  (\bibinfo{year}{2021}).
\newblock
\showeprint[arXiv]{2110.05286}
\urldef\tempurl%
\url{https://arxiv.org/abs/2110.05286}
\showURL{%
\tempurl}


\bibitem[Zhang et~al\mbox{.}(2021)]%
        {zhang2021cpm}
\bibfield{author}{\bibinfo{person}{Zhengyan Zhang}, \bibinfo{person}{Yuxian
  Gu}, \bibinfo{person}{Xu Han}, \bibinfo{person}{Shengqi Chen},
  \bibinfo{person}{Chaojun Xiao}, \bibinfo{person}{Zhenbo Sun},
  \bibinfo{person}{Yuan Yao}, \bibinfo{person}{Fanchao Qi},
  \bibinfo{person}{Jian Guan}, \bibinfo{person}{Pei Ke}, {et~al\mbox{.}}}
  \bibinfo{year}{2021}\natexlab{}.
\newblock \showarticletitle{Cpm-2: Large-scale cost-effective pre-trained
  language models}.
\newblock \bibinfo{journal}{\emph{AI Open}}  \bibinfo{volume}{2}
  (\bibinfo{year}{2021}), \bibinfo{pages}{216--224}.
\newblock


\bibitem[Zhao et~al\mbox{.}(2021)]%
        {DBLP:journals/ijccc/ZhaoYLGL21}
\bibfield{author}{\bibinfo{person}{Fengda Zhao}, \bibinfo{person}{Zhikai Yang},
  \bibinfo{person}{Xianshan Li}, \bibinfo{person}{Dingding Guo}, {and}
  \bibinfo{person}{Haitao Li}.} \bibinfo{year}{2021}\natexlab{}.
\newblock \showarticletitle{Extract Executable Action Sequences from Natural
  Language Instructions Based on {DQN} for Medical Service Robots}.
\newblock \bibinfo{journal}{\emph{Int. J. Comput. Commun. Control}}
  \bibinfo{volume}{16}, \bibinfo{number}{2} (\bibinfo{year}{2021}).
\newblock
\urldef\tempurl%
\url{https://doi.org/10.15837/ijccc.2021.2.4115}
\showURL{%
\tempurl}


\bibitem[Zhuo and Kambhampati(2013)]%
        {AMAN}
\bibfield{author}{\bibinfo{person}{Hankz~Hankui Zhuo} {and}
  \bibinfo{person}{Subbarao Kambhampati}.} \bibinfo{year}{2013}\natexlab{}.
\newblock \showarticletitle{Action-Model Acquisition from Noisy Plan Traces}.
  In \bibinfo{booktitle}{\emph{{IJCAI}}}. \bibinfo{pages}{2444--2450}.
\newblock


\bibitem[Zhuo and Kambhampati(2017)]%
        {DBLP:journals/ai/ZhuoK17}
\bibfield{author}{\bibinfo{person}{Hankz~Hankui Zhuo} {and}
  \bibinfo{person}{Subbarao Kambhampati}.} \bibinfo{year}{2017}\natexlab{}.
\newblock \showarticletitle{Model-lite planning: Case-based vs. model-based
  approaches}.
\newblock \bibinfo{journal}{\emph{Artif. Intell.}}  \bibinfo{volume}{246}
  (\bibinfo{year}{2017}), \bibinfo{pages}{1--21}.
\newblock
\urldef\tempurl%
\url{https://doi.org/10.1016/j.artint.2017.01.004}
\showURL{%
\tempurl}


\bibitem[Zhuo et~al\mbox{.}(2014)]%
        {DBLP:journals/ai/ZhuoM014}
\bibfield{author}{\bibinfo{person}{Hankz~Hankui Zhuo},
  \bibinfo{person}{H{\'{e}}ctor Mu{\~{n}}oz{-}Avila}, {and}
  \bibinfo{person}{Qiang Yang}.} \bibinfo{year}{2014}\natexlab{}.
\newblock \showarticletitle{Learning hierarchical task network domains from
  partially observed plan traces}.
\newblock \bibinfo{journal}{\emph{Artif. Intell.}}  \bibinfo{volume}{212}
  (\bibinfo{year}{2014}), \bibinfo{pages}{134--157}.
\newblock


\bibitem[Zhuo and Yang(2014)]%
        {DBLP:journals/ai/Zhuo014}
\bibfield{author}{\bibinfo{person}{Hankz~Hankui Zhuo} {and}
  \bibinfo{person}{Qiang Yang}.} \bibinfo{year}{2014}\natexlab{}.
\newblock \showarticletitle{Action-model acquisition for planning via transfer
  learning}.
\newblock \bibinfo{journal}{\emph{Artif. Intell.}}  \bibinfo{volume}{212}
  (\bibinfo{year}{2014}), \bibinfo{pages}{80--103}.
\newblock
\urldef\tempurl%
\url{https://doi.org/10.1016/j.artint.2014.03.004}
\showURL{%
\tempurl}


\bibitem[Zhuo et~al\mbox{.}(2020)]%
        {DBLP:journals/tist/ZhuoZKT20}
\bibfield{author}{\bibinfo{person}{Hankz~Hankui Zhuo}, \bibinfo{person}{Yantian
  Zha}, \bibinfo{person}{Subbarao Kambhampati}, {and} \bibinfo{person}{Xin
  Tian}.} \bibinfo{year}{2020}\natexlab{}.
\newblock \showarticletitle{Discovering Underlying Plans Based on Shallow
  Models}.
\newblock \bibinfo{journal}{\emph{{ACM} Trans. Intell. Syst. Technol.}}
  \bibinfo{volume}{11}, \bibinfo{number}{2} (\bibinfo{year}{2020}),
  \bibinfo{pages}{18:1--18:30}.
\newblock
\urldef\tempurl%
\url{https://doi.org/10.1145/3368270}
\showDOI{\tempurl}


\end{thebibliography}

\end{document}